\pdfoutput=1
\documentclass{article}

    \usepackage[nonatbib,final]{neurips_2022}

\usepackage[utf8]{inputenc} % allow utf-8 input
\usepackage[T1]{fontenc}    % use 8-bit T1 fonts
\usepackage{hyperref}       % hyperlinks
\usepackage{url}            % simple URL typesetting
\usepackage{booktabs}       % professional-quality tables
\usepackage{amsfonts}       % blackboard math symbols
\usepackage{nicefrac}       % compact symbols for 1/2, etc.
\usepackage{microtype}      % microtypography
\usepackage{xcolor}         % colors
\usepackage{amsmath}
\usepackage{amssymb}
\usepackage{graphicx}
\usepackage[linesnumbered]{algorithm2e}
\usepackage{qtree}
\usepackage[utf8]{inputenc}
\usepackage{ragged2e}
\usepackage{hyperref}

\title{Structural Kernel Search via Bayesian Optimization and Symbolical Optimal Transport}

\author{%
  Matthias Bitzer
    \quad
  Mona Meister
  \quad Christoph Zimmer \\
 	Bosch Center for Artificial Intelligence, Renningen, Germany\\
  \texttt{\{matthias.bitzer3,mona.meister,christoph.zimmer\}@de.bosch.com}
}

\begin{document}

\maketitle

\begin{abstract}
Despite recent advances in automated machine learning, model selection is still a complex and computationally intensive process. For Gaussian processes (GPs), selecting the kernel is a crucial task, often done manually by the expert. Additionally, evaluating the model selection criteria for Gaussian processes typically scales cubically in the sample size, rendering kernel search particularly computationally expensive. We propose a novel, efficient search method through a general, structured kernel space. Previous methods solved this task via Bayesian optimization and relied on measuring the distance between GP's directly in function space to construct a kernel-kernel. We present an alternative approach by defining a kernel-kernel over the symbolic representation of the statistical hypothesis that is associated with a kernel. We empirically show that this leads to a computationally more efficient way of searching through a discrete kernel space. 
\end{abstract}

\section{Introduction}
In many real-work applications of machine learning, tuning the hyperparameters or selecting the machine learning method itself is a crucial part of the workflow. It is often done by experts and data-scientist. However, the number of possible methods and models is constantly growing, and it is becoming increasingly important to automatically select the right model for the task at hand. Bayesian optimization (BO) is a prominent method that can be used for model selection and hyperparameter tuning. It can handle black-box oracles with expensive function evaluations, which are two characteristics often encountered when doing model selection \cite{PracticalBO}. Important applications in this context are choosing the hyperparameters in the training process of deep neural networks (DNN) \cite{PracticalBO}, or dealing with discrete and structured problems like choosing the architecture of DNN's \cite{NASviaOT,NASviaWassersteinOT}.

Gaussian processes (GP) are another important model-class. They are often utilized as surrogate models in BO \cite{PracticalBO}, for time-series and statistical modeling \cite{AutomaticStatistician,CKS} or in active learning loops \cite{SafeALTimeSeries,SchreiterSafeAL}. The properties of GP's are mainly governed by its kernel that specifies the assumptions made on the underlying function. Choosing the right kernel is therefore a crucial part of applying GP's and is often done by the expert. Recent work \cite{BOMS} treated the kernel selection as a black-box optimization problem and used Bayesian optimization to solve it. This allowed searching over a highly structured, discrete space of kernels. However, their proposed kernel-kernel measures the distance between two GP's directly in function space, which is a computationally expensive task itself. This makes the method difficult to apply for the frequent scenarios where the evaluation of the model selection criteria requires only a medium amount of time.

We propose measuring the distance between two kernels via their symbolical representation of their associated statistical hypothesis. We utilize the highly general kernel-grammar, presented in \cite{CKS}, as underlying kernel space, where each kernel is build from base kernels and operators, like e.g.
\begin{align*}
\mathrm{LIN}+((\mathrm{SE}\times\mathrm{PER})+ \mathrm{SE})
\end{align*}
forming effectively a description of the statistical hypothesis that is modeled by the GP. Our main idea is to build a distance over these symbolical descriptions, rather than measuring the distance between two GP's directly in function space. We employ optimal transport principles, known from neural-architecture search (see \cite{NASviaWassersteinOT},\cite{NASviaOT}), to build a pseudo-distance between two hypotheses descriptions and use it to construct a kernel-kernel, which is subsequently utilized in the BO loop. We will show that the induced kernel search method is more efficient, in terms of number of function evaluations and computational time, compared to alternative kernel search methods over discrete search spaces.

The main challenge we encountered is the quantification of dissimilarity between two symbolical representations of kernels. We use the tree representation of each symbolical description and apply optimal transport distances over tree features. Subsequently, we empirically show that the deduced GP over GP's provides a well-behaved meta-model and show its advantages for kernel search. In summary, our contributions are:
\begin{enumerate}
	\item We construct a pseudo-distance over GP's that acts over the symbolical representations of the underlying statistical hypothesis.
	\item We use the pseudo-distance to construct a novel "kernel-kernel" and build it in a  BO loop to do model selection for GP's. 
	\item We empirically show that our meta-GP model is well-behaved and that we outperform previous methods in kernel search over discrete kernel spaces.
\end{enumerate}

\paragraph{Related Work:} There is considerable existing work on constructing flexible kernels and learning their hyperparameters \cite{MultipleKernelLearning,WilsonExtrapolate,FastKernelLearningWilson,NKN,DKL}. In these kinds of works, the structural form of the kernel is predefined and the free parameters of the kernel are optimized, often via marginal likelihood maximization. While being able to efficiently finding the hyperparameter due to the differentiability of the marginal likelihood, one still needs to predefine the structural form of the kernel in the first place. This is a hard task as one need to decide if long-range correlation or nonstationarity should be considered or if dimensions are ignored or not. Our method is build to automatically select the structure of the kernel. Some of the mentioned methods \cite{WilsonExtrapolate,FastKernelLearningWilson} are able to approximate any stationary kernel via Bochner's Theorem and therefore consider a broad kernel space themselves. However, even elementary statistical hypotheses require nonstationary kernels, such as linear trends modeled by the linear kernel. Our search space is not restricted to stationary kernels.

We consider a highly general, discrete kernel space that is induced by the kernel grammar \cite{CKS}. Recent work \cite{BOMS} used BO to search through this space via a kernel that measures similarity in function space. We also utilize BO but employ a fundamentally different principle of measuring the distance, which is computationally more efficient. We dedicate Section \ref{hellingerkernelkernel} to a more precise comparison to the method of \cite{BOMS}. Additionally, the original kernel grammar paper \cite{CKS} suggested greedy search for searching through the kernel grammar. Furthermore, \cite{TreeGEP} employed a genetic algorithm based on cross-over mutations. We empirically compare against all approaches.

In the area of BO over structured spaces, our method is most similar to the neural-architecture search (NAS) procedures presented in \cite{NASviaWassersteinOT,NASviaOT} who use optimal transport distances over features extracted from the graph-representation of neural networks. Compared to these methods we use OT principles to do model selection for GP's which is a fundamentally different task.

\section{Background and Set-up}
Our main task is efficient model selection for Gaussian processes. In order to provide background information, we give a small introduction to GP's and model selection for GP's. Subsequently, we present a review of the kernel-grammar \cite{CKS} and show how we use it for our approach. We show how a kernel can be represented via a symbolical description. In Section \ref{sec:kernel_kernel}, we will present how we use the symbolical description to construct a kernel over kernels and how we use it in the BO loop.

\subsection{Gaussian Processes}
A Gaussian process is a distribution over functions $f:\mathcal{X}\to \mathbb{R}$ over a given input space $\mathcal{X}$ which is fully characterized via the covariance/kernel function $k(x,x')=\mathbf{Cov}(f(x),f(x'))$ and the mean function $\mu(x):=\mathbb{E}[f(x)]$. We therefore can write as shorthand notation $f\sim \mathcal{GP}(\mu(\cdot),k(\cdot,\cdot))$.
The kernel can be interpreted as a similarity measure between two elements of the input space, with the GP assigning higher correlations to function values whose inputs are more similar according to the kernel. Furthermore, the kernel governs the main assumptions on the modeled function such as smoothness, periodicity or long-range correlations and therefore provides the inductive-bias of the GP. An important property of GP's is that they are not restricted to euclidean input spaces $\mathcal{X}\subset \mathbb{R}^{d}$, but can also be defined on highly structured spaces like trees and graphs, a property we will later use to define a GP over GP's.

While our method might also be used for model selection in classification, we consider from now on Gaussian processes regression. For regression, a dataset $\mathcal{D}=(\mathbf{X},\mathbf{y})$ with $\mathbf{X}=\{x_{1},\dots,x_{N}\}\subset \mathcal{X}$ and $\mathbf{y}=(y_{1},\dots,y_{N})^{\intercal}\in \mathbb{R}^{N}$  is given, where we suppose that $f\sim \mathcal{GP}(\mu(\cdot),k(\cdot,\cdot))$ and $y_{i}=f(x_{i})+\epsilon_{i}$ with $\epsilon_{i}\overset{i.i.d}{\sim} \mathcal{N}(0,\sigma^{2})$. Given the observed data $\mathcal{D}$ the posterior distribution $f|\mathcal{D}$ is again a GP with mean and covariance functions
\begin{equation*}
\begin{aligned}
\label{eq:post_gp}
\mu_{\mathcal{D}}(x)&=\mu(x)+\mathbf{k}(x)^{\intercal}(\mathbf{K}+\sigma^{2}I)^{-1}(\mathbf{y}-\mu(\mathbf{X})),\\
k_{\mathcal{D}}(x,y)&=k(x,y)-\mathbf{k}(x)^{\intercal}(\mathbf{K}+\sigma^{2}I)^{-1}\mathbf{k}(y)
\end{aligned}
\end{equation*}
with $\mathbf{K}=[k(x_{m},x_{l})]_{m,l=1}^{N}$ and $\mathbf{k}(x)=[k(x,x_{1}),\dots,k(x,x_{N})]^{\intercal}$ (see \cite{3569}). Probabilistic predictions can be done via the resulting predictive distribution $p(f^{*}|x^{*},\mathcal{D})=\mathcal{N}(\mu_{\mathcal{D}}(x^{*}),k_{\mathcal{D}}(x^{*},x^{*}))$.

\subsection{Model selection for GP's}
Typically, the kernel $k_{\theta}$ comes with a set of parameters $\theta$ that can be learned via maximization of the marginal likelihood $p(\mathbf{y}|\mathbf{X},\theta,\sigma^{2})=\mathcal{N}(\mathbf{y};\mu(\mathbf{X}),k_{\theta}(\mathbf{X},\mathbf{X})+\sigma^{2}\mathbf{I})$ or via maximum a posteriori (MAP) estimation, in case the parameters $\theta$ are equipped with a prior $p(\theta)$.
 This procedure is sometimes also called model selection, as one selects the hyperparameters of the kernel given the data (see \cite{3569}). However, we consider selecting the \textit{structural form} of the kernel itself. The structural form of the kernel determines the statistical hypothesis that is assumed to be true for the data-generating process. Intuitively, the kernel is similar to the architecture in deep neural networks, which induces an inductive bias. 

Our goal is to do model selection over a discrete, possibly infinite space of kernels  $\mathbb{K}:=\{k_{1},k_{2},\dots\}$. As each kernel comes with its own parameters, we are actually dealing with a space of kernel families. Thus, when mentioning a kernel $k$ we associate it with its whole family over parameters $\{k_{\theta}|\theta \in \Theta\}$. Once a kernel is selected, predictions are done with learned kernel parameters (that usually are a by-product of calculating the model selection criteria). The parameters $\theta$ are potentially equipped with a prior $p(\theta)$ depending on the selection criteria. As mean function, we always utilize the zero mean function $\mu(x):=0$ in case $\mathbf{y}$ is centered and a constant mean function otherwise. This is a common choice in GP regression. Given some model selection criteria $g:\mathbb{K} \to \mathbb{R}$ our task is solving $k^{*} = \arg \max_{k\in\mathbb{K}} g(k|\mathcal{D})$. While our method is not restricted to a specific model selection criteria, we focus on the model evidence $p(\mathbf{y}|\mathbf{X},k)$, which is a well-known selection criteria for probabilistic models (see \cite{BOMS,occamsrazor,mackay92}). Given a prior on the kernel parameters $p(\theta)$ and the likelihood variance $p(\sigma^{2})$, the log-model evidence of the marginalized GP is given as
\begin{align*}
g(k|\mathcal{D})=\mathrm{log}~ p(\mathbf{y}|\mathbf{X},k)= \mathrm{log}\int p(\mathbf{y}|\mathbf{X},\theta,\sigma^{2},k) p(\sigma^{2})p(\theta|k) d\theta d\sigma^{2}.
\end{align*}
This quantity can be approximated via Laplace approximation of $p(\theta,\sigma^{2}|\mathcal{D})$ (see Appendix \ref{section:method_details} for details). Computing this approximation includes performing a MAP estimation of the GP parameters. Thus, once the log evidence has been computed, learned kernel hyperparameters $\theta_{\mathrm{MAP}}$ are automatically provided. Performing the MAP estimation scales cubically in the data set size $N$, which renders model selection for GP's computationally intense.

\subsection{Kernel Grammar}
The kernel grammar introduced by \cite{CKS} specifies a highly general search space over kernels. The grammar is based on the observation that kernels are closed under addition and multiplication, meaning, for kernels $k_{1}(x,x')$ and $k_{2}(x,x')$ also $k_{1}(x,x') + k_{2}(x,x')$ and $k_{1}(x,x') \times k_{2}(x,x')$ are kernels. Given some base kernels, such as the squared exponential kernel $\textrm{SE}$, the linear kernel $\textrm{LIN}$, the periodic kernel $\textrm{PER}$ or the rational quadratic kernel $\textrm{RQ}$, different statistical hypotheses can be generated via addition and multiplication. For example, 
\begin{align*}
\textrm{LIN}+\textrm{PER} \times \textrm{SE}
\end{align*}
describes a linear trend with a locally periodic component, such that it might be a useful hypothesis for time-series applications. For multidimensional data the base kernels are applied to single dimensions, denoted e.g. with $\mathrm{SE}_{i}$ for the squared exponential kernel defined on dimension $i$. The grammar contains many popular hypotheses such as the ARD-RBF kernel that can be expressed via multiplication over the dimensions $ \prod_{i=1}^{d}\textrm{SE}_{i}$ as well as additive models $\sum_{i=1}^{d}\textrm{SE}_{i}$ or polynomials of order $m$ with $\prod_{j=1}^{m}\textrm{LIN}$.

The specification of the complete search space is given as a grammar, where a base kernel is denoted as $\mathcal{B}$ and a subexpression is denoted with $\mathcal{S}$. For example, in the expression $\textrm{LIN}+(\textrm{PER} \times \textrm{SE})$ the expression $\textrm{PER} \times \textrm{SE}$ is a subexpression. Starting with all base kernels, the search space is defined as all kernels that can be reached via the following operations:
\begin{enumerate}
	\item Add a base kernel to a subexpression:
	$
	\mathcal{S} \rightarrow \mathcal{S} + \mathcal{B}
	$
	\item Multiply a subexpression with a base kernel:
	$
	\mathcal{S} \rightarrow \mathcal{S} \times \mathcal{B}
	$
	\item Exchange a base kernel with another base kernel:
	$
	\mathcal{B} \rightarrow \mathcal{B}'
	$
	
\end{enumerate}
Starting from the set of base kernels, these operations can lead to all algebraic expressions, showing the expressiveness of the grammar \cite{CKS}.
The authors of \cite{CKS} suggest using greedy search to search through the grammar, where the kernel with the highest value for the selection criteria is selected and expanded with all possible operations. After the selection criteria is calculated on the neighbors, the search progresses to the next stage by expanding the neighbors of the best kernel found.

We consider a generalized notion of the kernel grammar, where we consider a set of base kernels $\{\mathcal{B}_{1},\dots,\mathcal{B}_{r}\}, r\in \mathbb{N}$ and a set of operators $\{T_{1},\dots,T_{l}\}, l\in \mathbb{N}$ where $T_{j}:\mathcal{K}\times \mathcal{K} \to \mathcal{K}, j=1,\dots,l$ are closed operators on the space of all kernel functions $\mathcal{K}$. Examples are addition and multiplication, but also the change-point operator which was considered in \cite{AutomaticStatistician}. Thus, in general the grammar operations are
\begin{enumerate}
	\item Apply operator $T_{j}$ onto a subexpression and a base kernel\footnote{In case the operator $T_{j}$ is not symmetric we also add the operation $
		\mathcal{S} \rightarrow T_{j}(\mathcal{B},\mathcal{S}) 
		$. However, all operators considered in this work are symmetric.}:
	$
	\mathcal{S} \rightarrow T_{j}(\mathcal{S},\mathcal{B}) 
	$ 
	\item Exchange a base kernel with another base kernel:
	$
	\mathcal{B} \rightarrow \mathcal{B}'
	$
	
\end{enumerate}
\newtheorem{definition}{Definition}
The considered kernel space can be defined precisely in the following way:
\begin{definition}
	\label{kernel_space}
	For $r,l \in \mathbb{N}$, let $\{k_{1},\dots,k_{r}\}$ be a set of (base-) kernels (symbollically represented as $\{\mathcal{B}_{1},\dots,\mathcal{B}_{r}\}$) and $\{T_{1},\dots,T_{l}\}$ a set of operators with $T_{j}:\mathcal{K}\times \mathcal{K} \to \mathcal{K}, j=1,\dots,l$. Let 
	\begin{align*}
	L_{0}&:=\{k_{1},\dots,k_{r}\}\\
	L_{i}&:=\{T_{j}(k_{1},k_{2})~|~k_{1},k_{2}\in L_{i-1},j=1,\dots,l\} \cup L_{i-1}
	\end{align*}
	for $i=1,\dots,M$. We call $\tilde{\mathbb{K}}:= L_{M}$ the kernel-grammar generated kernel space with depth $M$.
\end{definition}

\subsection{Representation of Kernels}
The base kernels and their combination via the operators describe the structural assumptions of the final kernel and imply the assumptions that are made in function space. Our main hypothesis is that the symbolical representation of the kernel, given by the subexpressions $\mathcal{S}$ and base kernels, already contains sufficient information for model selection. Therefore, we will define a kernel-kernel over the symbolical representations and utilize it for Bayesian optimization. Concretely, each kernel $k\in \tilde{\mathbb{K}}$ in our described space can be written as a tree $\mathcal{T}$, for example:
\begin{equation*}
\mathrm{LIN}+((\mathrm{SE}+\mathrm{PER})\times \mathrm{SE}) ~~~\longleftrightarrow ~~~\vcenter{\hbox{ \Tree [.{\textbf{ADD}} {LIN} [.{\textbf{MULT}} [.{\textbf{ADD}} {SE} {PER} ] SE ] ]}}
\end{equation*}
Each operator $T_{j}$ and each base kernel $\mathcal{B}_{i}$ is represented by their respective name, where operators are the nodes of the tree and base kernels are the leafs. The way how the operators and base kernels are connected is represented through the tree structure. For a given expression tree $\mathcal{T}$, we denote the multiset of all subexpressions/subtrees as $\mathrm{Subtrees}(\mathcal{T})$. Furthermore, we consider paths to the leafs of the tree, for example, for the tree above one path to a leaf would be:
\begin{align*}
\mathbf{ADD} \longrightarrow \mathbf{MULT} \longrightarrow \mathbf{ADD} \longrightarrow \mathrm{PER}
\end{align*}
We denote the multiset of all paths in the tree $\mathcal{T}$ as $\mathrm{Paths}(\mathcal{T})$. Lastly, we also consider the multiset of all base kernels that exist in a given expression tree as $\mathrm{Base}(\mathcal{T})$. For each described multiset, we denote the number of occurrences of element $\mathcal{E}$ in the multiset as $n(\mathcal{E})$. When building the multisets we also account for two symmetries in the elements that can be applied if an operator is associative and commutative, which we elaborate further in Appendix \ref{section:method_details}.

Depending on the operators that are used, several trees can describe the same kernel $k\in\tilde{\mathbb{K}}$. For technical reasons, we denote with $f: \tilde{\mathbb{K}}\mapsto\mathcal{T}(\tilde{\mathbb{K}})$ a mapping that maps a given kernel $k\in \tilde{\mathbb{K}}$ to one tree $\mathcal{T}$ that induces this kernel, where $\mathcal{T}(\tilde{\mathbb{K}})$ denotes the set of all expression trees that can generate a kernel in $\tilde{\mathbb{K}}$ (see details in Appendix \ref{section:method_details}).  
\section{Kernel-Kernel}
\label{sec:kernel_kernel}
Our kernel-kernel will be defined via a pseudo-metric over the expression trees. Optimal transport (OT) principles have proven themselves to be effective in BO methods over structured spaces, as shown in \cite{NASviaOT} and \cite{NASviaWassersteinOT} for neural architecture search or in \cite{ChemBO} for BO over molecule structures. We follow this line of work and also rely on optimal transport, although we only use a simple ground metric to allow for closed-form computations. 
To allow OT metrics to be used, we summarize each expression tree $\mathcal{T}$ to discrete probability distributions of their building blocks $\mathrm{Base}(\mathcal{T})$, $\mathrm{Paths}(\mathcal{T})$ and $\mathrm{Subtrees}(\mathcal{T})$ via
\begin{align*}
\omega_{\mathrm{base}}:=\sum_{\mathcal{E}\in \mathrm{Base}(\mathcal{T})} \omega_{\mathcal{E}} \delta_{\mathcal{E}}~~,~~~~~~
\omega_{\mathrm{paths}}:=\sum_{\mathcal{E}\in \mathrm{Paths}(\mathcal{T})} \omega_{\mathcal{E}} \delta_{\mathcal{E}}~~,~~~~~~~
\omega_{\mathrm{subtrees}}:=\sum_{\mathcal{E}\in \mathrm{Subtree}(\mathcal{T})} \omega_{\mathcal{E}} \delta_{\mathcal{E}}~~,
\end{align*}
where $\delta_{\mathcal{E}}$ is the Dirac delta and $\omega_{\mathcal{E}}$ is the frequency of the element $\mathcal{E}$ in the respective multiset. For example, the frequency of expression $\mathrm{SE} \times \mathrm{PER}$ in the multiset $\mathrm{Subtree}(\mathcal{T})$ is calculated as
\begin{align*}
\omega_{\mathrm{SE} \times \mathrm{PER}}=\frac{n(\mathrm{SE} \times \mathrm{PER})}{|\mathrm{Subtree}(\mathcal{T})|}.
\end{align*}
Each probability distribution represents a different modeling aspect that is induced by a kernel $k\in\tilde{\mathbb{K}}$ and its corresponding expression tree $\mathcal{T}$:
\begin{enumerate}
	\item $\omega_{\mathrm{base}}$ specifies \textit{which} base kernels are present in $\mathcal{T}$, thus, which base assumptions in function space are included such as periodicity, linearity, local smoothness.
	\item $\omega_{\mathrm{paths}}$ specifies \textit{how} the base kernels in $\mathcal{T}$ are used, e.g. whether a periodic component is applied additively or multiplicatively.
	\item $\omega_{\mathrm{subtrees}}$ specifies the \textit{interaction} between the base kernels in $\mathcal{T}$, for example if $\mathcal{T}$ contains an addition of a linear and a periodic component or not.
\end{enumerate}
Our pseudo-metric between kernels $k_{1}$ and $k_{2}$ (and its associated trees $\mathcal{T}_{1}$ and $\mathcal{T}_{2}$) uses all three modeling aspects via the optimal transport distances between the respective discrete probability distributions $\omega_{\mathrm{base}}$, $\omega_{\mathrm{paths}}$ and $\omega_{\mathrm{subtrees}}$.

In general the OT distance with ground metric $\tilde{d}$ between two discrete probability distribution $\omega_{1}=\sum_{\mathcal{E}\in\Omega}\omega_{1,\mathcal{E}}\delta_{\mathcal{E}}$ and $\omega_{2}=\sum_{\mathcal{E}\in\Omega}\omega_{2,\mathcal{E}}\delta_{\mathcal{E}}$ over $\Omega$ is defined as 
\begin{equation}\label{OT}
W_{\tilde{d}}(\omega_{1},\omega_{2}) = \mathrm{inf}_{\pi \in \mathcal{R}(\omega_{1},\omega_{2})} \int_{\Omega \times \Omega} \tilde{d}(\mathcal{E},\mathcal{E}') \pi(d\mathcal{E},d\mathcal{E}'),
\end{equation}
where $\pi \in \mathcal{R}(\omega_{1},\omega_{2})$ is a combined probability distribution over $\Omega \times \Omega$ with marginal distribution $\pi(A\times \Omega)=\omega_{1}(A)$ and $\pi(\Omega \times B)=\omega_{2}(B)$ for Borel sets $A,B$. While for general ground metric $\tilde{d}:\Omega \times \Omega \to \mathbb{R}$ the optimization problem in $(\ref{OT})$ can not be solved in closed-form we use as ground metric $\tilde{d}(\mathcal{E},\mathcal{E}')=\mathbf{1}_{\mathcal{E}\neq \mathcal{E}'}$ which has as closed-form solution the total variation distance between $\omega_{1}$ and $\omega_{2}$ (see \cite{villani2008optimal}, p. 22), i.e.
\begin{align*}
W_{\tilde{d}}(\omega_{1},\omega_{2}) = \frac{1}{2}\sum_{\mathcal{E} \in \Omega} |\omega_{1,\mathcal{E}}-\omega_{2,\mathcal{E}}|.
\end{align*}
Utilizing this distance over the modeling assumptions $\omega_{\mathrm{base}}$, $\omega_{\mathrm{paths}}$ and $\omega_{\mathrm{subtrees}}$ allows for fast computation of the final pseudo-metric and for a proper positive semi definite (p.s.d) kernel-kernel in the end [see Appendix \ref{section:proofs}]. This is not guarantueed for general OT distances (see \cite{NASviaWassersteinOT}).
We define the final distance between two kernels $k_{1}$ and $k_{2}$ (and its associated trees $\mathcal{T}_{1}$ and $\mathcal{T}_{2}$) as a sum over the OT distances of their respective modeling components
\begin{align}
\label{dist1}
\begin{split}
d(\mathcal{T}_{1},\mathcal{T}_{2}):= &\alpha_{1}W_{\tilde{d}}(\omega_{1,\mathrm{base}},\omega_{2,\mathrm{base}})\\+&\alpha_{2}W_{\tilde{d}}(\omega_{1,\mathrm{paths}},\omega_{2,\mathrm{paths}})\\+&\alpha_{3}W_{\tilde{d}}(\omega_{1,\mathrm{subtrees}},\omega_{2,\mathrm{subtrees}}),
\end{split}
\end{align}
where $\alpha_{i}\ge 0$ and $\sum_{i}\alpha_{i}=1$ are weighting parameters that will later be learned automatically via marginal likelihood maximization for GPs.

In case the kernel grammar contains base kernels that act on single dimensions such as $\mathrm{SE}_{i}$, we define for each dimension $i=1,\dots,D$ an individual distribution over base kernels
\begin{align*}
\omega_{\mathrm{base}}^{(i)}:=\sum_{\mathcal{E}\in \mathrm{Base}(\mathcal{T},i)} \omega_{\mathcal{E}} \delta_{\mathcal{E}}~~,
\end{align*}
where $\mathrm{Base}(\mathcal{T},i)$ is the multiset of all base kernels in $\mathcal{T}$ defined on dimension $i$. We also include the empty-expression $\mathcal{E}_{\mathrm{NULL}}$ for which $\omega_{\mathcal{E}_{\mathrm{NULL}}}=1$ whenever no base kernel of dimension $i$ is contained in $\mathcal{T}$. $\omega_{\mathrm{base}}^{(i)}$ thus summarizes which base kernels are present that act on dimension $i$ or if the dimension is ignored. The distance in this case is defined as
\begin{align}
\label{dist2}
\begin{split}
d(\mathcal{T}_{1},\mathcal{T}_{2}):= &\alpha_{1}\sum_{i=1}^{D}W_{\tilde{d}}(\omega_{1,\mathrm{base}}^{(i)},\omega_{2,\mathrm{base}}^{(i)}) \\+&\alpha_{2}W_{\tilde{d}}(\omega_{1,\mathrm{paths}},\omega_{2,\mathrm{paths}}) \\+&\alpha_{3}W_{\tilde{d}}(\omega_{1,\mathrm{subtrees}},\omega_{2,\mathrm{subtrees}}).
\end{split}
\end{align}
The reason for this distinction is that two kernels $k_{1}$ and $k_{2}$, which have different active dimensions, are considered to be farther apart with this distance, making it easier to allow variable selection, which is an important aspect of model selection.

Our defined function $d(\mathcal{T}_{1},\mathcal{T}_{2})$ induces indeed a pseudo-metric in the kernel-grammar generated kernel space, as shown in the following proposition. In particular, it fulfills the triangle inequality.
\newtheorem{proposition}{Proposition}
\begin{proposition}
	\label{pseudo_metric}
	Let $\tilde{\mathbb{K}}$ be the kernel space generated by a kernel grammar. Let $f: \tilde{\mathbb{K}}\mapsto\mathcal{T}(\tilde{\mathbb{K}})$ be a mapping that maps a kernel $k\in \tilde{\mathbb{K}}$ to one of its expression trees $\mathcal{T}$. Then $\hat{d}(k_{1},k_{2}):=d(f(k_{1}),f(k_{2}))$ is a pseudo-metric on $\tilde{\mathbb{K}}$ where $d$ is given by (\ref{dist1}) or (\ref{dist2}) depending on the base kernels in $\tilde{\mathbb{K}}$.
\end{proposition}
Given the pseudo-metric, we are now able to define a kernel-kernel with
\begin{equation}
K_{SOT}(k_{1},k_{2}):=\sigma^{2}\mathrm{exp}\bigg(\frac{-\hat{d}(k_{1},k_{2})}{l^{2}}\bigg),
\end{equation}
which we call \textit{Symbolical-Optimal-Transport (SOT)} kernel-kernel. Here, $l$ denotes the lengthscale and $\sigma^{2}$ the variance of the kernel-kernel. Both parameters are learned in combination with the distance weights via marginal likelihood maximization (see Appendix \ref{section:method_details}).
\subsection{Bayesian Optimization for Model Selection}
We utilize the proposed kernel-kernel to do model selection for GP's via Bayesian optimization, which is a similar task to \cite{BOMS}. Compared to \cite{BOMS}, we use our proposed kernel-kernel, which is a fundamentally different and computationally more efficient way of measuring similarity in GP space.

Given a model selection criteria $g_{\mathcal{D}}: \tilde{\mathbb{K}} \to \mathbb{R}$ for a given dataset $\mathcal{D}$, we want to solve $k^{*} = \arg \max_{k\in \tilde{\mathbb{K}}} g(k|\mathcal{D})$ via Bayesian optimization. We thus define a surrogate GP model for $g(k|\mathcal{D})$ via
\begin{align*}
f \sim \mathcal{GP}(\mu_c(\cdot),K_{SOT}(\cdot,\cdot)),
\end{align*}
where $\mu_c(k)=c$ is the constant mean function. As BO acquisition function $a(k|\tilde{\mathcal{D}}_{t})$ we use Expected-Improvement (EI). Here, $\tilde{\mathcal{D}}_{t}$ denotes the set of already queried kernel-selection-criteria pairs $(k,g(k|\mathcal{D}))$ for which the meta-GP posterior $f|\tilde{\mathcal{D}}_{t}$ is calculated. Starting with an initial set of pairs $\tilde{\mathcal{D}}_{0}$ we follow the standard BO iterations where in each iteration the acquisition function is maximized $k_{t}=\arg \max_{k\in \tilde{\mathbb{K}}} a(k|\tilde{\mathcal{D}}_{t})$ given the current set of already evaluated kernel-selection-criteria pairs $\tilde{\mathcal{D}}_{t}$. Then the selection criteria is queried at the chosen kernel $g(k_{t}|\mathcal{D})$. A complete description of the method can be found in Algorithm \ref{alg:bo} in Appendix \ref{section:method_details}.

A crucial part in the BO cycle is the optimization of the acquisition function $\max_{k\in \tilde{\mathbb{K}}} a(k|\tilde{\mathcal{D}}_{t})$. While for Euclidean spaces gradient-based methods or grid-based methods could be used to solve that task, this is not an option for a structured space like the considered grammar-generated kernel space. We propose using an evolutionary algorithm to optimize the acquisition function, which can be seen in Algorithm \ref{alg:ea} in Appendix \ref{section:method_details}. While an evolutionary algorithm seems to be a computationally intense procedure that takes place in each BO iteration, we emphasize that the evaluation of the acquisition function is very efficient for our proposed method. The reason for the efficiency is that the evaluations of our kernel-kernel $K_{SOT}(k_{1},k_{2})$ at two kernels $k_{1},k_{2}\in\tilde{\mathbb{K}}$ is very cheap, compared for example to the method in \cite{BOMS} (see Appendix \ref{section:further_experiments} for computational time comparision).
\subsection{Comparision to Hellinger Kernel-Kernel}
\label{hellingerkernelkernel}
The work of \cite{BOMS} also uses BO for kernel selection but with a different principle of measuring the distance in GP space. In particular, they propose measuring the distance of two GP's $\mathcal{M}$ and $\mathcal{M}'$ via the induced prior distributions $p(\mathbf{f}|\mathbf{X},\mathcal{M})$ in function space evaluated on the design matrix $\mathbf{X}$ of the dataset. Conditioned on the parameters of the GP $\theta$, the distributions $p(\mathbf{f}|\mathbf{X},\mathcal{M},\theta)$ are Gaussian and they use the Hellinger distance over Gaussians as base distance between GP models $d(\mathcal{M}_{\theta},\mathcal{M}'_{\theta'}|\theta,\theta')$ given the hyperparameters. They finally take the expectation over the hyperparameter priors to construct their final distance. Thus, for each kernel-kernel evaluation $K(\mathcal{M},\mathcal{M}')$ the integral
\begin{align*}
d(\mathcal{M},\mathcal{M}')=\int \int d(\mathcal{M}_{\theta},\mathcal{M}'_{\theta'}|\theta,\theta') p(\theta)p(\theta') d\theta d\theta'
\end{align*}
 needs to be computed, where the integrand scales cubically in $|\mathbf{X}|$. Each kernel-kernel evaluation is therefore a computationally hard problem by itself. They side-pass this issue partly by selecting a subset of $\mathbf{X}$ as input locations and by using sample-based estimation of the integral. While this could make the method useful for oracles with very long run times, such as model selection for very large data sets, we see conceptual problems in the case of medium oracle run times. For our method, kernel-kernel evaluations $K(k_{1},k_{2})$ are very efficient as they scale only in the size of the expression tree and no integrals need to be computed.
 \begin{table}[t]
 	\centering
 	\caption{RMSE on test kernel-log-evidence pairs ($\alpha=0.05$)}
 	\label{rmse_table_space1}
 	\begin{tabular}{lllll}
 		\toprule
 		{Dataset} & Hellinger &kNN & Mean & SOT (ours)\\
 		\midrule
 		$\mathrm{Airfoil}$     &                                            0.1773 (0.004) &            0.2231 (0.013) &       0.3769 (0.002) &                                   \textbf{0.0936} (0.003) \\
 		$\mathrm{Airline}$ &                                            0.3569 (0.007) &            0.3813 (0.011) &       0.4013 (0.005) &                                   \textbf{0.3464} (0.011) \\
 		$\mathrm{LGBB}$       &                                            0.3772 (0.021) &            0.6795 (0.025) &       0.8519 (0.016) &                                   \textbf{0.2783} (0.008) \\
 		$\mathrm{Powerplant}$ &                                            0.1912 (0.012)&            0.2236 (0.011) &       0.2925 (0.009) & \textbf{0.0137} (0.005) \\
 		$\mathrm{Concrete}$ & 0.2489 (0.006) &            0.1806 (0.010) &       0.2912 (0.005) &                                   \textbf{0.0451} (0.002) \\
 		\bottomrule
 	\end{tabular}
 \end{table}
\section{Experiments}
\label{section:experiments}
In the following section, we show experimental results for our novel meta-GP model and kernel search method.
We evaluate our meta-model on a meta-regression task where we predict test kernel-log-evidence pairs based on training pairs. Secondly, we consider kernel search and compare it against the greedy method in \cite{CKS}, the evolutionary algorithm in \cite{TreeGEP} and against the BO method of \cite{BOMS}. The implementation of our method is available at  \url{https://github.com/boschresearch/bosot}.
\paragraph{Selection Criteria and Datasets:} In all our experiments, we consider the normalized log-model evidence $g(k|\mathcal{D})=\mathrm{log}~ p(\mathbf{y}|\mathbf{X},k)/|\mathcal{D}|$ as model selection criteria, as also done in \cite{BOMS}. We always use the Laplace approximation to calculate the log-model evidence, where we use 10 repeats to do MAP estimation of the kernel parameters. Furthermore, we consider the following publicly available datasets: $\mathrm{Airline},\mathrm{LGBB},\mathrm{Airfoil},\mathrm{Powerplant},\mathrm{Concrete}$. $\mathrm{Airline}$ is a one dimensional time-series dataset, $\mathrm{LGBB}$ is a two-dimensional dataset with low observation noise. $\mathrm{Powerplant},\mathrm{Airfoil}$ and $\mathrm{Concrete}$ are four, five, and eight dimensional, whereas $\mathrm{Powerplant}$ exhibits higher observation noise. We use medium-sized training sets - 100 and 150 datapoints for $\mathrm{Airline}$ and $\mathrm{LGBB}$ and 500 datapoints for the other three datasets. All training sets are uniformly drawn from the full datasets. The outputs are normalized and the inputs are scaled to be in the unit interval. Further details can be found in Appendix \ref{section:experimental_details}.
\begin{figure}[t]
	\centering
	\includegraphics[width=0.99\linewidth]{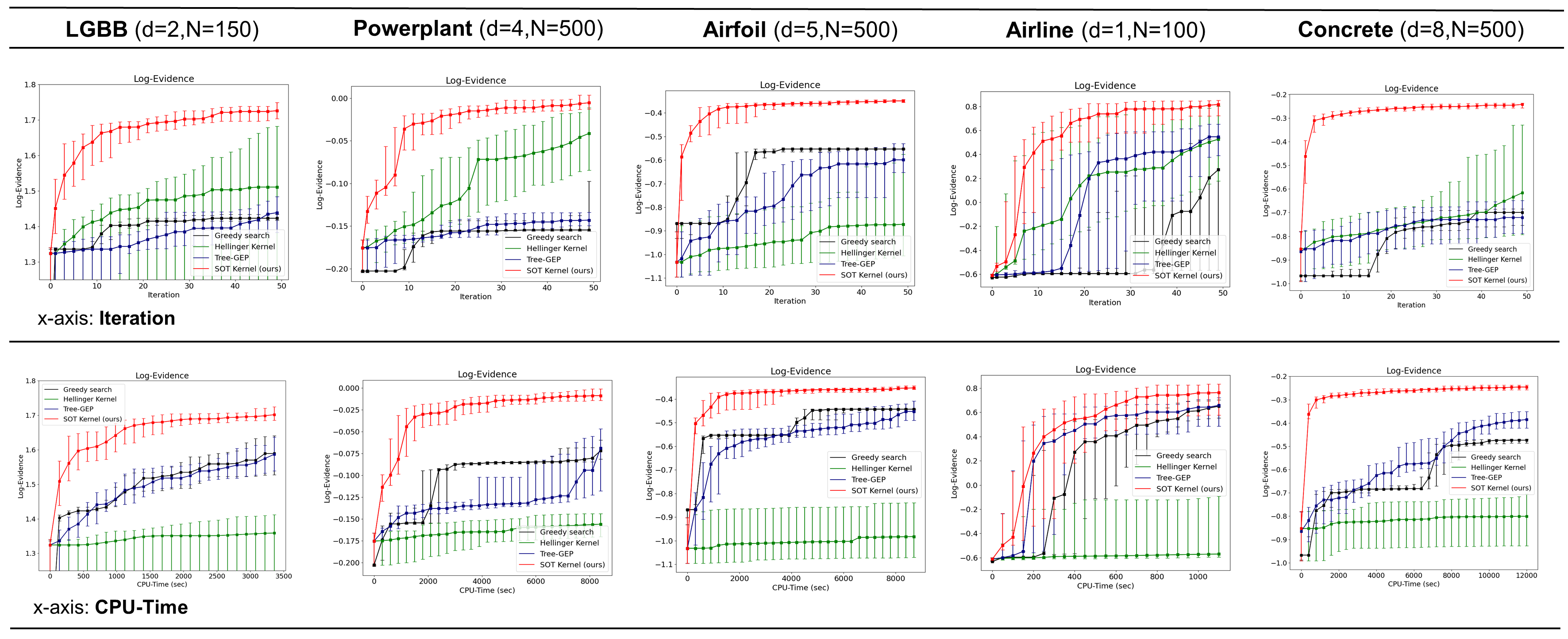}
	\caption{Plots for the model selection task over number of evaluated models (top) and CPU-time (bottom) showing the normalized log-evidence of the best model found up to this iteration/time point.}
	\label{fig:mainresults}
\end{figure}
\paragraph{Search Spaces:}
We consider two search spaces. The first consists of the base kernels  $\mathrm{SE}_{i},\mathrm{LIN}_{i},\mathrm{PER}_{i},\mathrm{RQ}_{i}$ and the operators $+$ and $\times$. This is the space considered in \cite{CKS} for time series and low-dimensional datasets. The second search space uses as base kernels $\mathrm{SE}_{i}$ and $\mathrm{RQ}_{i}$ and also $+$ and $\times$ as operators and was considered in \cite{CKS} for higher dimensional base datasets. We consider the first space for $\mathrm{Airline}$ and $\mathrm{LGBB}$ and the second for $\mathrm{Powerplant}, \mathrm{Airfoil}$, and $\mathrm{Concrete}$. The hyperparameter priors for the base kernels can be found in the  Appendix \ref{section:method_details}.
\paragraph{Prediction of Selection Criteria:} We evaluate our meta-GP model on a meta-regression task by predicting test kernel-log-evidence pairs based on training pairs. The quality of predictions in kernel space might also be an indicator for good performance in BO for kernel search. Given the base dataset $\mathcal{D}$ we create a training set $\tilde{\mathcal{D}}_{\mathrm{train}}=\{(k_{i},g(k_{i}|\mathcal{D}))| k_{i}\in \tilde{\mathbb{K}}_{\mathrm{train}}\subset \tilde{\mathbb{K}}, i=1,\dots,n_\mathrm{train}\}$ and a test set $\tilde{\mathcal{D}}_{\mathrm{test}}=\{(k_{i},g(k_{i}|\mathcal{D}))| k_{i}\in \tilde{\mathbb{K}}_{\mathrm{test}}\subset \tilde{\mathbb{K}}, i=1,\dots,n_\mathrm{test}\}$ - each containing 500 kernel-log-evidence pairs. We generate the train and tests sets exactly as in \cite{BOMS}, where we create one set $\tilde{\mathbb{K}}_{\mathrm{complete}}$ by first initializing it with all base kernels and iteratively pick one kernel of the current set, apply one random operation of the kernel grammar and add the resulting kernel to the current set and repeat this process until we have $n_\mathrm{train}+n_\mathrm{test}$ kernels in $\tilde{\mathbb{K}}_{\mathrm{complete}}$. We then divide the set uniformly into $\tilde{\mathbb{K}}_{\mathrm{train}}$ and $\tilde{\mathbb{K}}_{\mathrm{test}}$. We compare our model to the mean-predictor as baseline, which just predicts the mean of the train set at each test point and a kNN predictor based on the kernel-grammar operations [see Appendix \ref{section:experimental_details} or \cite{BOMS}]. Furthermore, we compare against the meta-GP model of \cite{BOMS}. We report root mean squared error (RMSE) scores on the test sets in Table \ref{rmse_table_space1}. It can be observed that our method leads to more precise predictions on all four meta prediction tasks, indicating that the symbolical representations already contain sufficient information to predict log-evidence values. 
\paragraph{Model Selection - Setup:} Concerning kernel search, we compare against the BO method that employs the Hellinger kernel-kernel \cite{BOMS}, against greedy search \cite{CKS} and against the evolutionary algorithm presented in \cite{TreeGEP}, referred to as TreeGEP. Both BO methods run for 50 iterations and the kernel-kernel hyperparameters are updated in each iteration via marginal likelihood maximization. Our method applied the evolutionary Algorithm \ref{alg:ea} (see Appendix \ref{section:method_details}) to optimize its acquisition function, using a population size of 100. As it is computationally unfeasible to apply the same kind of acquisition function optimization for the Hellinger kernel-kernel we use their method of optimizing the acquisition function where an active set of kernels is updated in each iteration. Both BO methods and TreeGEP start with the same set of initial kernels, for which we apply two random grammar operations for each base kernel. Greedy search by design needs to start from the empty kernel. We give it a head start by the number of initial datapoints (see Appendix \ref{section:experimental_details}). For each dataset, we display results from 30 independent runs with different seeds, namely medians and quartiles of the log-evidence score over iterations and CPU-time in Figure \ref{fig:mainresults}. The implementation of both BO methods is based on \textit{GPflow} \cite{GPflow2017}. Further experimental details and parameter settings for all methods can be found in Appendix \ref{section:experimental_details}. 
\paragraph{Model Selection - Results:} As shown in Figure \ref{fig:mainresults}, we outperform all methods in terms of performance over number of model evaluations. This is not surprising against the two heuristics, as they are not optimized towards keeping the number of model evaluations low. However, it is notable that we are more sample-efficient compared to \cite{BOMS}, who solve a much harder problem for computing their kernel-kernel. In terms of performance over CPU-time, we outperform all methods on $\mathrm{Airfoil},\mathrm{Powerplant},\mathrm{LGBB}$ and $\mathrm{Concrete}$ significantly. On $\mathrm{Airline}$, the advantage over the heuristics is smaller - the reason is the lower oracle time - which benefits the heuristics that do not need to optimize the acquisition function. The reason for the poor performance of the Hellinger kernel-kernel in terms of CPU-time is the high ratio of acquisition function optimization to oracle time, which was as high as $50:1$ in our experiments (see detailed numbers in Appendix \ref{section:experimental_details}).
\paragraph{Test Performance:} When optimizing a model selection criteria, one expects that this also materializes in a better test performance. In Table \ref{rmseTest} we therefore show the predictive negative log-likelihood (NLL) scores on a held out test-set of the selected models at the final time stamp. We observe that the advantage in the model selection value gets transferred to the test performance.
\paragraph{Further Experiments:} In Appendix \ref{section:further_experiments} we include further investigations on the behavior of our method on simulated data coming from a  ground-truth kernel. Furthermore, we include a comparison of the selected kernel to the standard RBF kernel and against Functional Kernel Learning (FKL) \cite{FKL}.

\begin{table}[t]
	\centering
	\caption{Predictive negative log-likelihood values on test set at final time stamp. Values are marked bold if they are not significantly different from the best value according to a two-sample t-test ($\alpha=0.05$).}
	\label{rmseTest}
	\begin{tabular}{lcccl}
		\toprule
		Dataset & Greedy & Hellinger & Tree-GEP  & SOT (ours)\\
		\midrule
		$\mathrm{Airline}$ & -0.4042 (0.615) &   ~0.3368 (0.207) & \textbf{-0.5594} (0.580) & \textbf{-0.7015} (0.471)\\
		$\mathrm{LGBB}$  & -0.7528 (0.661) &    \textbf{-0.8787} (0.854) & \textbf{-1.0701} (0.532) & \textbf{-0.9325} (0.492)  \\
		$\mathrm{Powerplant}$ &  -0.0053 (0.054) &   ~0.0580 (0.037) & -0.0241 (0.057) & \textbf{-0.0661} (0.032)\\
		$\mathrm{Airfoil}$  & ~\textbf{0.0837} (0.026) &   ~0.9080 (0.205) &  ~0.1826 (0.118) &  ~\textbf{0.1006} (0.090)\\
		$\mathrm{Concrete}$ & ~0.3254 (0.019) &   ~0.6633 (0.207) &  ~\textbf{0.2812} (0.074) &  ~\textbf{0.2872} (0.044) \\
		\bottomrule
	\end{tabular}

\end{table}

\section{Limitations}
\label{section:limitations}
In case the dataset size is very small, greedy search or evolutionary algorithms might have an advantage in terms of computational time as the optimization of the acquisition function outweighs the computation of the model selection criteria. Thus, in these instances our method might not show a strong benefit. Furthermore, the kernel grammar is often used for downstream applications that use the selected hypothesis for interpretation, such as \cite{AutomaticStatistician} who build an automatic natural language description of the selected hypothesis. Depending on the dataset and how many steps are employed in the acquisition function optimization, relatively large hypotheses might be found as optimal. This might render the interpretation difficult. However, we show in Appendix \ref{section:further_experiments} example configurations of our algorithm that can be used to get smaller, well interpretable hypothesis.

\section{Conclusion}
We presented a novel way of doing BO for model selection of GP's by measuring the distance between two GP's via the symbolical description of the underlying statistical hypothesis. The main contribution is the deduced pseudo-metric over kernels and the resulting SOT kernel-kernel. We show that our approach leads to a more efficient way of searching through a discrete kernel space compared to other BO methods and search heuristics.

%%%%%%%%%%%%%%%%%%%%%%%%%%%%%%%%%%%%%%%%%%%%%%%%%%%%%%%%%%%%
\bibliography{lib}
\bibliographystyle{abbrv}

\section*{Checklist}

\begin{enumerate}

	\item For all authors...
	\begin{enumerate}
		\item Do the main claims made in the abstract and introduction accurately reflect the paper's contributions and scope?
		\answerYes{We clarify our main contributions in the end of the introduction and provide the mentioned method in Section \ref{sec:kernel_kernel} and the mentioned experimental results in Section \ref{section:experiments}.}
		\item Did you describe the limitations of your work?
		\answerYes{In Section \ref{section:limitations}}
		\item Did you discuss any potential negative societal impacts of your work?
		\answerNo{No obvious direct negative societal impacts.}
		\item Have you read the ethics review guidelines and ensured that your paper conforms to them?
		\answerYes{}
	\end{enumerate}

	\item If you are including theoretical results...
	\begin{enumerate}
		\item Did you state the full set of assumptions of all theoretical results?
		\answerYes{Given in Definition \ref{kernel_space}, Proposition \ref{pseudo_metric} and in Section \ref{section:proofs}}
		\item Did you include complete proofs of all theoretical results?
		\answerYes{In Section \ref{section:proofs}}
	\end{enumerate}

	\item If you ran experiments...
	\begin{enumerate}
		\item Did you include the code, data, and instructions needed to reproduce the main experimental results (either in the supplemental material or as a URL)?
		\answerNA{We will provide code as soon as clearance of code is completed.}
		\item Did you specify all the training details (e.g., data splits, hyperparameters, how they were chosen)?
		\answerYes{Either in Section \ref{section:experiments} or Appendix \ref{section:experimental_details} }
		\item Did you report error bars (e.g., with respect to the random seed after running experiments multiple times)?
		\answerYes{We report either standard deviations or quartiles.}
		\item Did you include the total amount of compute and the type of resources used (e.g., type of GPUs, internal cluster, or cloud provider)?
		\answerYes{We analyze the search performance over CPU-time.}
	\end{enumerate}

	\item If you are using existing assets (e.g., code, data, models) or curating/releasing new assets...
	\begin{enumerate}
		\item If your work uses existing assets, did you cite the creators?
		\answerYes{Yes we cite the GPflow related work, which is the main python framework we are using. }
		\item Did you mention the license of the assets?
		\answerNo{}
		\item Did you include any new assets either in the supplemental material or as a URL?
		\answerYes{Code is available at \url{https://github.com/boschresearch/bosot}.}
		\item Did you discuss whether and how consent was obtained from people whose data you're using/curating?
		\answerNA{}
		\item Did you discuss whether the data you are using/curating contains personally identifiable information or offensive content?
		\answerNA{}
	\end{enumerate}

	\item If you used crowdsourcing or conducted research with human subjects...
	\begin{enumerate}
		\item Did you include the full text of instructions given to participants and screenshots, if applicable?
		\answerNA{}
		\item Did you describe any potential participant risks, with links to Institutional Review Board (IRB) approvals, if applicable?
		\answerNA{}
		\item Did you include the estimated hourly wage paid to participants and the total amount spent on participant compensation?
		\answerNA{}
	\end{enumerate}

\end{enumerate}

\newpage
\appendix

\section{Method Details}
\label{section:method_details}
In the following section, we specify further details of the proposed method. We show the symmetries we employ in the multisets and give details on the approximation of the log-evidence. Furthermore, we describe the base kernels and their parameter priors, the marginal likelihood maximization of the kernel-kernel hyperparameters and how the acquisition function optimization is done. 
\subsection{Symmetries in the Multisets}
\label{section:symmetries}
Our method can be used with general operators $T:\mathcal{K}\times\mathcal{K} \to \mathcal{K}$ in kernel space. However, depending on the concrete operators, one might also incorporate properties that leave the expression/subexpressions unchanged. In particular, addition and multiplication are commutative and associative. For commutative operators two trees/subtrees describe the same kernel if one rotates the nodes under a commutative operator, e.g.
\begin{equation*}
\mathcal{S}_{1}=\vcenter{\hbox{ \Tree [.{\textbf{ADD}} {LIN} [.{\textbf{ADD}} [.{\textbf{MULT}} {PER} {SE} ] SE ] ]}}~~~~~~~~~~\mathcal{S}_{2}=\vcenter{\hbox{ \Tree [.{\textbf{ADD}} {LIN} [.{\textbf{ADD}} SE [.{\textbf{MULT}} {PER} {SE} ] ] ]}}
\end{equation*}
This symmetry is automatically considered in the multisets $\mathrm{Base}(\mathcal{T})$ and  $\mathrm{Path}(\mathcal{T})$ as the base kernels and paths stay unchanged when rotating two nodes. For $\mathrm{Subtree}(\mathcal{T})$, we account explicitly  for that symmetry via hashing functions that are invariant to rotation of subtrees. This allows counting tree $\mathcal{S}_{1}$ and tree $\mathcal{S}_{2}$ as an identical subtree $\tilde{\mathcal{S}}$ in the multiset $\mathrm{Subtree}(\mathcal{T})$.

For the multiset $\mathrm{Path}(\mathcal{T})$, we furthermore consider a symmetry that exists in case an operator is associative and commutative. In this case, one can exchange the nodes of two consecutive operators of the same kind without changing the expression, e.g.
\begin{equation*}
\vcenter{\hbox{ \Tree [.{\textbf{ADD}} {LIN} [.{\textbf{ADD}} [.{\textbf{MULT}} {PER} {SE} ] SE ] ]}}~~\Leftrightarrow~~\vcenter{\hbox{ \Tree [.{\textbf{ADD}} {SE} [.{\textbf{ADD}} [.{\textbf{MULT}} {PER} {SE} ] LIN ] ]}} ~~\Leftrightarrow~~ \vcenter{\hbox{ \Tree [.{\textbf{ADD}} [.{\textbf{MULT}} {PER} {SE} ] [.{\textbf{ADD}} {SE} {LIN} ] ]}} .
\end{equation*}
We incorporate this symmetry into $\mathrm{Path}(\mathcal{T})$ by only considering the smallest path to that base kernel that exist in an equivalent expression (equivalent under this symmetry). This is realized in $\mathrm{Path}(\mathcal{T})$ by counting identical operators in a row along a path only as one, e.g. the following paths would be considered the same:
\begin{align*}
&\mathbf{ADD} \longrightarrow \mathbf{ADD} \longrightarrow \mathbf{MULT} \longrightarrow \mathrm{PER}~,\\
&\mathbf{ADD}  \longrightarrow \mathbf{MULT} \longrightarrow \mathrm{PER}~.
\end{align*}
While one might also integrate more symmetries that stem from, e.g. the distributive property of the multiplication, we found that these two symmetries are particularly easy and efficient to implement. The symmetries are also not restricted to multiplication or addition. The first symmetry can be used whenever $T$ is associative, thus, $T(k_{1},k_{2})=T(k_{2},k_{1})$ and the second whenever $T$ is associative and commutative, thus, $T(k_{1},T(k_{2},k_{3}))=T(k_{2},T(k_{1},k_{3}))=T(T(k_{1},k_{3}),k_{2})$. The motivation of considering symmetries at all is that two trees $\mathcal{T}_{1}$ and $\mathcal{T}_{2}$ are considered more similar in case they share the same structure (with respect to the symmetry) and, therefore, similar kernels can be detected more easily.

\subsection{Mapping from Kernels to Trees}
As we already observed in the previous section, two different trees $\mathcal{T}_{1}$ and $\mathcal{T}_{2}$ might describe the same kernel $k$. For technical reasons, we therefore consider in Proposition \ref{pseudo_metric} a mapping $f:\mathbb{K} \to \mathcal{T}(\tilde{\mathbb{K}})$ that maps a kernel always to the same tree. Concretely, this is done such that the pseudo metric is defined in kernel space rather than in tree space. When using addition and multiplication, one could check if $\mathcal{T}_{1}$ and $\mathcal{T}_{2}$ describe the same expression via recursive hashes that follow the same rules as addition and multiplication. One could use this check to implement such a function $f:\mathbb{K} \to \mathcal{T}(\tilde{\mathbb{K}})$. However, in our experiments we directly deal with the trees (the BO and the acquisition function algorithm directly act on the trees anyway) and ignore filtering out potential collisions of two trees - we did not observe any downsides of doing that.

\subsection{Approximation of Log-Model-Evidence}
\label{section:laplace}
In all our experiments, we use the normalized log-model-evidence $g(k|\mathcal{D})=\mathrm{log}~ p(\mathbf{y}|\mathbf{X},k)/|\mathcal{D}|$ as selection criteria. Similar to \cite{BOMS} we use the Laplace approximation to approximate the log-evidence, which is (see \cite{BOMS})
\begin{align*}
\mathrm{log}~p(\mathbf{y}|\mathbf{X},k) \approx \mathrm{log}~p(\mathbf{y}|\mathbf{X},k,\hat{\gamma}) + \mathrm{log}~ p(\hat{\gamma})-\frac{1}{2}\mathrm{log}~\mathrm{det}~\Sigma^{-1} +\frac{d}{2}~\mathrm{log}~2 \pi,
\end{align*}
where $\gamma=(\theta,\sigma^{2}) \in \mathbb{R}^{d}$ are the parameters of the kernel and the likelihood variance, $d\in \mathbb{N}$ is the number of parameters, $\hat{\gamma}$ denotes the MAP estimate of $\gamma$, and $\Sigma^{-1} = - \nabla^{2} \mathrm{log}~p(\gamma|\mathcal{D},k)|_{\gamma=\hat{\gamma}}$. Creating the MAP estimate of $\gamma$ scales cubically in the dataset size $|\mathcal{D}|$ in each optimization step. We use the LBFGS optimizer to create the MAP estimate. As the loss function is non-convex, we make 10 restarts with random initialization of the initial parameters.

\subsection{Base Kernels and Priors on Parameters:}
Here, we specify the base kernels that are used, including their priors on the parameters. We chose the parameter priors such that broad priors in function space are induced (Here, we assume that the datasets contain normalized outputs and inputs scaled to the unit interval).  All base kernels are defined on $\mathbb{R}$ and are applied on dimension $i$ if this is indicated by the base kernel symbol, e.g. $\mathrm{SE}_{i}$. We consider the following base kernels:
\begin{enumerate} 
\item Squared Exponential $\mathrm{SE}$:
\begin{align*}
k(x,x')=\sigma^{2}\mathrm{exp}\bigg(-\frac{1}{2} \frac{(x-x')^{2}}{l^{2}} \bigg)
\end{align*}
with $l \sim \mathrm{Gamma}(2.0,2.0)$ and $\sigma^{2} \sim \mathrm{Gamma}(2.0,3.0)$,
\item Periodic $\mathrm{PER}$:
\begin{align*}
k(x,x')=\sigma^{2}\mathrm{exp}\bigg(-\frac{1}{2} \frac{\mathrm{sin}^{2}(\pi|x-x'|/p)}{l^{2}} \bigg)
\end{align*}
with $l \sim \mathrm{Gamma}(2.0,2.0)$, $\sigma^{2} \sim \mathrm{Gamma}(2.0,3.0)$, and $p \sim \mathrm{Gamma}(2.0,2.0)$,
\item Linear $\mathrm{LIN}$:
\begin{align*}
k(x,x')=\sigma^{2}x\,y + \sigma_{c}^{2}
\end{align*}
with $\sigma^{2} \sim \mathrm{Gamma}(2.0,3.0)$ and $\sigma_{c}^{2} \sim \mathrm{Gamma}(2.0,3.0)$,
\item Rational Quadratic $\mathrm{RQ}$:
\begin{align*}
k(x,x')=\sigma^{2}\bigg(1+\frac{(x-x')^{2}}{2\alpha l^{2}} \bigg)^{-\alpha}
\end{align*}
with $l \sim \mathrm{Gamma}(2.0,2.0)$, $\sigma^{2} \sim \mathrm{Gamma}(2.0,3.0)$, and $\alpha \sim \mathrm{Gamma}(2.0,2.0)$.
\end{enumerate}
\begin{algorithm}[t]
	\label{alg:bo}
	
	\SetKwFunction{BomsSOT}{BOKernelSearch}
	\SetKwFunction{EvolutionaryAlg}{EvolutionaryAlg}
	
	\SetKwProg{Fn}{Function}{:}{}
	\Fn{\BomsSOT{$\mathcal{D}$,$T$,$L$,$\mathrm{n_{initial}}$}}{
		$\tilde{\mathcal{D}}_{0}=\mathrm{GetInitialDataset(\mathcal{D},\mathrm{n_{initial}})}$\\
		\For{$t=0,\dots,T-1$}{
			Fit Meta GP model $f \sim \mathcal{GP}(\mu_c(\cdot),K_{SOT}(\cdot,\cdot))$ to $\tilde{\mathcal{D}}_{t}$\\
			$k_{t} \leftarrow $ \EvolutionaryAlg{$a(\cdot|\tilde{\mathcal{D}}_{t})$,$L$}\\
			Query model selection criteria $g_{t} \leftarrow g(k_{t}|\mathcal{D})$\\
			$\tilde{\mathcal{D}}_{t+1}=\tilde{\mathcal{D}}_{t}\cup \{(k_{t},g_{t})\}$
			
		}
		$t^{*}\leftarrow \arg \max_{t=0,\dots,T-1} g_{t}$
		
		\Return{$k_{t^{*}}$}
	}
	\vspace{0.3cm}
	\caption{BO for kernel search via SOT kernel-kernels.}
\end{algorithm}

\begin{algorithm}[b]
	\label{alg:ea}
	
	\SetKwFunction{EvolutionaryAlg}{EvolutionaryAlg}
	
	\SetKwProg{Fn}{Function}{:}{}
	\Fn{\EvolutionaryAlg{$a(\cdot|\tilde{\mathcal{D}}_{t})$,$L$}}{
		$\mathrm{n_{survive}}=\frac{\mathrm{n_{population}}}{(\mathrm{n_{offspring}}+1)}$\\
		$\mathcal{K}_{0}=\mathrm{GetInitialKernels(n_{population})}$\\
		\For{$l=0,\dots,L-1$}{
			$\mathrm{fitness}_{l}=a(\mathcal{K}_{l}|\tilde{\mathcal{D}}_{t})$\\
			$\mathcal{K}_{l}^{\mathrm{selected}}=\mathrm{Select}(\mathcal{K}_{l},\mathrm{fitness}_{l},\mathrm{n_{survive}})$\\
			$\mathcal{K}_{l}^{\mathrm{offspring}}=\emptyset$\\
			\For{$k~\mathbf{in}~\mathcal{K}_{l}^{\mathrm{selected}}$}{
				$\mathcal{K}_{l}^{\mathrm{offspring}} \leftarrow \mathcal{K}_{l}^{\mathrm{offspring}} \cup \mathrm{ApplyGrammarOps}(k,\mathrm{n_{offspring}})$}
			$\mathcal{K}_{l+1}=\mathcal{K}_{l}^{\mathrm{selected}}\cup \mathcal{K}_{l}^{\mathrm{offspring}}$
			
		}
		\Return{$\arg \max(a(\mathcal{K}_{L}|\tilde{\mathcal{D}}_{t}))$}
	}
	\vspace{0.3cm}
	\caption{Evolutionary algorithm for BO kernel search.}
\end{algorithm}
\subsection{Marginal Likelihood Maximization of Kernel-Kernel Parameters}
\label{section:marginal_likelihood}
Doing GP regression with the meta GP model $f \sim \mathcal{GP}(\mu_c(\cdot),K_{SOT}(\cdot,\cdot))$ involves fitting some parameters, namely, the constant $c \in \mathbb{R}$ in the mean function, the kernel-kernel variance $\sigma^{2}$, the kernel-kernel lengthscale $l^{2}$, the distance weights $\alpha_{1},\alpha_{2},\alpha_{3}$, and the likelihood variance denoted by $\sigma_{g}^{2}$. The distance weights are reparameterized with $\alpha_{i}:=\frac{\sigma(\tilde{\alpha}_{i})}{\sum_{j=1}^{3}\sigma(\tilde{\alpha}_{j})},\tilde{\alpha}_{j} \in \mathbb{R}$, where $\sigma(\cdot)$ is the standard sigmoid function. Thus, $\alpha_{1}+\alpha_{2}+\alpha_{3}=1$ and $\alpha_{i}\ge 0$. For the other parameters, we use standard \textit{GPflow} bijectors to transform them to their domain of definition. Let $\theta$ denote all kernel-kernel parameters and let $\tilde{\mathcal{D}}=\{(k_{j},g(k_{j}|\mathcal{D}))| j=1,\dots,R\}$ denote the kernel-selection-criteria pairs to which the meta-GP model is to be fitted. We fit the parameter via maximization of the log-marginal likelihood
\begin{align}
(\theta^{*},\sigma_{g}^{*},c^{*}) = \arg \max_{\theta, \sigma_{g},c} \mathrm{log}\,\mathcal{N}(\mathbf{g};\mu_{c}(\mathbf{X}_{\mathrm{kernels}}),K_{SOT,\theta}(\mathbf{X}_{\mathrm{kernels}},\mathbf{X}_{\mathrm{kernels}})+\sigma_{g}^{2}\mathbf{I})
\end{align}
with $\mathbf{g}=[g(k_{1}|\mathcal{D}),\dots,g(k_{R}|\mathcal{D})]^{\intercal}$ and $\mathbf{X}_{\mathrm{kernels}}=\{k_{1},\dots,k_{R}\}$.

\subsection{BO Algorithm and Acquisition Function Optimization}
\label{section:EAandBOalg}
In Algorithm \ref{alg:bo}, we show the BO steps for kernel search. First, we draw an initial dataset of kernel-selection-criteria pairs. Our base setting applies two random grammar operations from each base kernel. The meta-GP parameters are fitted in each BO optimization. The acquisition function is optimized via the evolutionary algorithm in \ref{alg:ea}. This algorithm searches in the hypotheses space from small to big hypotheses. Given an initial population of kernels, we calculate the acquisition function on all kernels in the population. Then the $n_{\mathrm{survive}}$ best kernels in the population survive. Each selected kernel gets $n_{\mathrm{offspring}}$ children by generating $n_{\mathrm{offspring}}$ new kernels via applying one random grammar operation. The new population is formed via the selected kernel combined with its offspring. $L$ is the number of iterations in the evolutionary algorithm. In combination with the initial population it determines how many base kernels the hypotheses can contain maximally. In each iteration, the number of base kernels in the best hypothesis can maximally grow by one.

\section{Example Calculation of SOT Kernel-Kernel}
\label{section:example}
In this section, we give a small example calculation of our proposed kernel-kernel. We calculate our proposed pseudo-metric $d_{SOT}(k_{1},k_{2})$ for the two kernels $k_{1}$ and $k_{2}$ with the following expression trees
\begin{equation*}
\mathcal{T}_{1}=\vcenter{\hbox{ \Tree [.{\textbf{MULT}} {LIN} [.{\textbf{ADD}} [.{\textbf{MULT}} {PER} {SE} ] SE ] ]}}~~~~\text{and}~~~~\mathcal{T}_{2}=\vcenter{\hbox{ \Tree [.{\textbf{MULT}} [.{\textbf{ADD}} LIN SE ] [.{\textbf{ADD}} [.{\textbf{MULT}} {PER} {LIN} ] SE ] ]}} .
\end{equation*}
In the first step, we extract the tree features and create the multisets. Here, we use the notation $\{(\mathcal{E}_{1};n(\mathcal{E}_{1})),(\mathcal{E}_{2};n(\mathcal{E}_{2})),\dots\}$ to denote the existence of an element $\mathcal{E}$ in the multiset as well as the cardinality of the element. We obtain the following multisets:
\begin{align*}
&\mathrm{Base}(\mathcal{T}_{1})=\{(\mathrm{LIN};1),(\mathrm{SE};2),(\mathrm{PER};1)\},\\
&\mathrm{Base}(\mathcal{T}_{2})=\{(\mathrm{LIN};2),(\mathrm{SE};2),(\mathrm{PER};1)\},\\
&\mathrm{Path}(\mathcal{T}_{1})=\{(\mathbf{MULT} \longrightarrow \mathbf{ADD} \longrightarrow \mathbf{MULT} \longrightarrow \mathrm{PER};1),\\
&~~~~~~~~~~~~~~~~~~~~~~~(\mathbf{MULT} \longrightarrow \mathbf{ADD} \longrightarrow \mathbf{MULT} \longrightarrow \mathrm{SE};1),\\
&~~~~~~~~~~~~~~~~~~~~~~~(\mathbf{MULT} \longrightarrow \mathbf{ADD} \longrightarrow \mathrm{SE};1)\\
&~~~~~~~~~~~~~~~~~~~~~~~(\mathbf{MULT} \longrightarrow \mathrm{LIN};1)\},\\
&\mathrm{Path}(\mathcal{T}_{2})=\{(\mathbf{MULT} \longrightarrow \mathbf{ADD} \longrightarrow \mathbf{MULT} \longrightarrow \mathrm{PER};1),\\
&~~~~~~~~~~~~~~~~~~~~~~~(\mathbf{MULT} \longrightarrow \mathbf{ADD} \longrightarrow \mathbf{MULT} \longrightarrow \mathrm{LIN};1),\\
&~~~~~~~~~~~~~~~~~~~~~~~(\mathbf{MULT} \longrightarrow \mathbf{ADD} \longrightarrow \mathrm{SE};2)\\
&~~~~~~~~~~~~~~~~~~~~~~~(\mathbf{MULT} \longrightarrow \mathbf{ADD} \longrightarrow \mathrm{LIN};1)\},\\
&\mathrm{Subtree}(\mathcal{T}_{1})=\{(\mathcal{T}_{1};1),\left(\vcenter{\hbox{ \Tree [.{\textbf{ADD}} [.{\textbf{MULT}} {PER} {SE} ] SE ] }};1\right),\left(\vcenter{\hbox{ \Tree [.{\textbf{MULT}} {PER} {SE} ]}};1\right),(\mathrm{LIN};1),(\mathrm{SE};2),(\mathrm{PER};1)\},\\
&\mathrm{Subtree}(\mathcal{T}_{2})=\{(\mathcal{T}_{2};1),\left(\vcenter{\hbox{ \Tree [.{\textbf{ADD}} [.{\textbf{MULT}} {PER} {LIN} ] SE ] }};1\right),\left(\vcenter{\hbox{ \Tree [.{\textbf{MULT}} {PER} {LIN} ]}};1\right),\left(\vcenter{\hbox{ \Tree [.{\textbf{ADD}} {LIN} {SE} ]}};1\right),(\mathrm{LIN};2),\\
&~~~~~~~~~~~~~~~~~~~~~~~(\mathrm{SE};2),(\mathrm{PER};1)\}.
\end{align*}
Next, we build the probability vector for each feature multiset, for example, for the base kernels:
\begin{align*}
&\omega_{1,\mathrm{base}}=\frac{1}{4}\delta_{\mathrm{LIN}}+\frac{1}{2}\delta_{\mathrm{SE}}+\frac{1}{4}\delta_{\mathrm{PER}}\,, \\
& \omega_{1,\mathrm{base}}=\frac{2}{5}\delta_{\mathrm{LIN}}+\frac{2}{5}\delta_{\mathrm{SE}}+\frac{1}{5}\delta_{\mathrm{PER}}\,,
\end{align*}
and calculate the total variation distance between $\omega_{1,\mathrm{base}}$ and $\omega_{2,\mathrm{base}}$:
\begin{align*}
W_{\tilde{d}}(\omega_{1,\mathrm{base}},\omega_{2,\mathrm{base}}) =\frac{1}{2}\bigg(\bigg|\frac{1}{4}-\frac{2}{5}\bigg|+\bigg|\frac{1}{2}-\frac{2}{5}\bigg|+\bigg|\frac{1}{4}-\frac{1}{5}\bigg|\bigg)=\frac{3}{20}.
\end{align*}
For the paths, the total variation distance results to:
\begin{align*}
W_{\tilde{d}}(\omega_{1,\mathrm{path}},\omega_{2,\mathrm{path}}) =\frac{1}{2}\bigg(\bigg|\frac{1}{4}-\frac{1}{5}\bigg|+\bigg|\frac{1}{4}-0\bigg|+\bigg|\frac{1}{4}-\frac{2}{5}\bigg|+\bigg|\frac{1}{4}-0\bigg|+\bigg|0-\frac{1}{5}\bigg|+\bigg|0-\frac{1}{5}\bigg|\bigg)=\frac{11}{20}.
\end{align*}
For the subtrees, the total variation distance results to:
\begin{align*}
W_{\tilde{d}}(\omega_{1,\mathrm{path}},\omega_{2,\mathrm{path}}) &=\frac{1}{2}\bigg(\bigg|\frac{1}{7}-0\bigg|+\bigg|\frac{1}{7}-0\bigg|+\bigg|\frac{1}{7}-0\bigg|+\bigg|\frac{1}{7}-\frac{2}{9}\bigg|+\bigg|\frac{2}{7}-\frac{2}{9}\bigg|+\bigg|\frac{1}{7}-\frac{1}{9}\bigg|\\
&+\bigg|0-\frac{1}{9}\bigg|+\bigg|0-\frac{1}{9}\bigg|+\bigg|0-\frac{1}{9}\bigg|+\bigg|0-\frac{1}{9}\bigg|\bigg)=\frac{11}{21}.
\end{align*}
The complete distance given the weights is then
\begin{align*}
d(k_{1},k_{2})=\alpha_{1}\frac{3}{20}+\alpha_{2}\frac{11}{20}+\alpha_{3}\frac{11}{21}
\end{align*}
and
\begin{align*}
K_{SOT}(k_{1},k_{2}):=\sigma^{2}\mathrm{exp}\bigg(\frac{-\tilde{d}(k_{1},k_{2})}{l^{2}}\bigg).
\end{align*}
The parameters are learned via marginal likelihood maximization and are dependent on the dataset.

\begin{algorithm}[t]
	\label{alg:tree}
	
	\SetKwFunction{GetKernel}{GetKernel}
	
	\SetKwProg{Fn}{Function}{:}{}
	\Fn{\GetKernel{$\mathcal{T}$}}{
		\eIf{Root of $\mathcal{T}$ is leaf}{
	\Return base kernel $\mathcal{B}$ associated with leaf}{ 
\Return $T(\GetKernel(\mathcal{T}_{L}),\GetKernel(\mathcal{T}_{R}))$ where $T$ is the operator associated with the root of $\mathcal{T}$ and $\mathcal{T}_{L}$ and $\mathcal{T}_{R}$ are the left and right subtree below the root of $\mathcal{T}$.
}}

	\vspace{0.3cm}
	\caption{Get a kernel from an expression tree.}
\end{algorithm}
\section{Technical Details and Proofs}
\label{section:proofs}
When referring to a kernel $k$ from the kernel-grammar generated kernel space $\tilde{\mathbb{K}}$ such as writing $k\in \tilde{\mathbb{K}}$, we actually refer to the associated kernel family over its parameters $\{k_{\theta}|\theta \in \Theta\}$. Applying an operator $T$ onto two kernels $k_{1}$ and $k_{2}$ then results in another kernel family $\{T(k_{1,\theta_{1}},k_{2,\theta_{2}})|\theta_{1} \in \Theta_{1},\theta_{2}\in \Theta_{2}\}$ and we refer with the notation $T(k_{1},k_{2})$ to this kernel family with new parameter space $\Theta_{1}\times \Theta_{2}$. In case the same kernel family $\{\tilde{k}_{\theta}|\theta \in \Theta\}$ appears twice in an operator the parameters are not shared, e.g. $T(\tilde{k},\tilde{k})$ corresponds to the family $\{T(\tilde{k}_{\theta_{1}},\tilde{k}_{\theta_{2}})|\theta_{1} \in \Theta,\theta_{2}\in \Theta\}$ with new parameter space $\Theta \times \Theta$.
\begin{definition}
	We call a binary tree $\mathcal{T}$ whose nodes are associated with the operators $\{T_{1},\dots,T_{l}\}$ and whose leafs are associated with the base kernels $\{\mathcal{B}_{1},\dots,\mathcal{B}_{r}\}$ an expression tree of a kernel $k$ if $k$ is constructed by applying the operators recursively onto the leafs, meaning $k$ is the result of Algorithm \ref{alg:tree}.
\end{definition}
Any kernel $k$ in a kernel-grammar generated kernel space has an expression tree $\mathcal{T}$ via the construction of $\tilde{\mathbb{K}}$. We further denote the set of all trees that can generate a kernel in $\tilde{\mathbb{K}}$ as $\mathcal{T}(\tilde{\mathbb{K}}):=\{\mathcal{T}|\exists k\in \tilde{\mathbb{K}} : k~\text{is result of}~ \mathrm{GetKernel}(\mathcal{T}) \}$.
\paragraph{Proof of Proposition 1:} W.l.o.g. we consider $\tilde{\mathbb{K}}$ without separate base kernels for dimensions.
We denote by $g_{i}:\mathcal{T}(\tilde{\mathbb{K}})\to [0,1]^{L_{i}}$ and $i\in\{\mathrm{base},\mathrm{path},\mathrm{subtree}\}$ the mappings from expression trees to the probability vectors $\omega_{\mathrm{base}},\omega_{\mathrm{paths}},\omega_{\mathrm{subtrees}}$, where $L_{i}$ denotes the number of different elements of the respective type (e.g. number of different paths to the leafs for expression trees of depth $M$). Then
\begin{align*}
\hat{d}(k_{1},k_{2})=d(f(k_{1}),f(k_{2}))&=\frac{\alpha_{1}}{2}\Vert g_{1}(f(k_{1}))- g_{1}(f(k_{2})) \Vert_{1}\\&+\frac{\alpha_{2}}{2}\Vert g_{2}(f(k_{1}))- g_{2}(f(k_{2})) \Vert_{1}\\&+\frac{\alpha_{3}}{2}\Vert g_{3}(f(k_{1}))- g_{3}(f(k_{2})) \Vert_{1}
\end{align*}
is a pseudo metric as chaining of a metric with a mapping results in a pseudo metric, and the positive-weighted sum of pseudo metrics still is a pseudo metric. $\blacksquare$

\begin{proposition}
	Let $\tilde{\mathbb{K}}$ be the kernel space generated by a kernel grammar. Then 
	\begin{equation}
	K_{SOT}(k_{1},k_{2}):=\sigma^{2}\mathrm{exp}\bigg(\frac{-\hat{d}(k_{1},k_{2})}{l^{2}}\bigg)
	\end{equation} is a proper p.s.d. kernel over $\tilde{\mathbb{K}}$.
\end{proposition}
\paragraph{Proof:} As in the proof of Proposition 1 we can write $\hat{d}(k_{1},k_{2})$ as a weighted sum of Manhatten metrics, which leads to:
	\begin{align*}
K_{SOT}(k_{1},k_{2})&:=\sigma^{2}\mathrm{exp}\bigg(\frac{-\hat{d}(k_{1},k_{2})}{l^{2}}\bigg)\\
&=\sigma^{2}\prod_{i=1}^{3}\mathrm{exp}(-\frac{\alpha_{i}}{2\,l^{2}}\Vert g_{i}(f(k_{1}))- g_{i}(f(k_{2})) \Vert_{1}).
\end{align*} 
Thus, $K_{SOT}(k_{1},k_{2})$ can be written as a product of  Ornstein-Uhlenbeck kernels chained with functions $g_{i}\circ f$. Kernels chained with arbitrary mappings are kernels and products of kernels are kernels (see \cite{kernel_methods_pattern}, Proposition 3.22). Thus, $K_{SOT}(k_{1},k_{2})$ is a proper (p.s.d.) kernel. $\blacksquare$

\section{Experimental Details}
\label{section:experimental_details}
In this section, we give further details on parameter configuration and implementation details of the different methods.

\paragraph{Datasets:} All datasets are publically available. $\mathrm{Powerplant},\mathrm{Airfoil}$ and $\mathrm{Concrete}$ are UCI regression datasets (\url{https://archive.ics.uci.edu/ml/datasets.php}). For $\mathrm{LGBB}$ and $\mathrm{Airline}$, we describe how to access the datasets in the accompanying code at \url{https://github.com/boschresearch/bosot}.

\paragraph{SOT Kernel-Kernel (Our method):} We use Algorithm \ref{alg:ea} to optimize the acquisition function. Our base setting uses a population size of 100 and $n_{\mathrm{offspring}}=4$. The number of selected kernels is chosen such that the population stays constant. We choose 10 optimization steps for the bigger search spaces LGBB, Powerplant as well as Airfoil and 6 steps for the smaller search space Airline. We chose this number depending on the number of base kernels in the search space.

\paragraph{Hellinger Kernel-Kernel:}
For the Hellinger Kernel-Kernel \cite{BOMS}, we use their principle of optimizing the acquisition function, where an active set of kernels is kept in memory over the iterations. In each iteration, 15 random walks are performed in the kernel grammar, where the walk length is drawn randomly from a geometric distribution with $p=\frac{1}{3}$. These kernels are added to the active set. Furthermore, the neighbors of the best kernel found so far are added to the active set (we limit this set to 50 neighbors per iteration, since for large expressions there may be several hundred neighbors, making it computationally almost infeasible to evaluate the kernel-kernel on all of them). The active set is limited to 600 kernels, where only the ones are kept with the highest acquisition function value. We initialize the active set with random walks in the grammar, with one random walk of length 5 from each base kernel (this is an alternative to their computationally very expensive version of using all kernels two edges apart from the base kernels). We stick to their base settings of using 100 hyperparameter samples and a subset of the design matrix $\mathbf{X}$ of size of 20 inside the Hellinger distance calculation. We cache all distance calculations $d(\mathcal{M},\mathcal{M}')$ over the iterations, thus, recalculation of the Hellinger distance for kernels in the active set or the current dataset is very cheap. On the other hand, calculating the distance that involves a new kernel, such as a new neighbor of the currently best kernel, is very expensive. Given the distance matrix between kernels, optimization of the kernel-kernel hyperparameters is cheap again. 

\paragraph{Greedy Search:}
The greedy method in \cite{CKS} starts with the empty kernel, then evaluates all base kernels. It picks the best performing base kernel and determines its immediate neighbors via expanding the base kernels with all possible grammar operations. Then the neighbors are evaluated (we pick the order in which the neighbors are evaluated at random - for each seed a different order). Once all neighbors have been evaluated, the best kernel is determined and the process repeats. We give greedy search a head start of $n_{initial}$ log-evidence evaluations in Figure \ref{fig:mainresults}, meaning log-evidence values are shown once greedy search has evaluated as many kernels as the other methods have in their initial dataset.

\paragraph{TreeGEP:}
For the evolutionary algorithm in \cite{TreeGEP}, we use their base settings in the paper, which is a population size of 200, a reproduction rate of 0.1, and a probability of mutation vs. cross-over of 50\% each. They don't specify the size of the mutation subtrees and the tournament fraction in the tournament selection. Here, we make a reasonable choice and generate mutation subtrees of size four and use a tournament fraction of 0.1.

\begin{figure}[t]
	\centering
	\includegraphics[width=0.99\linewidth]{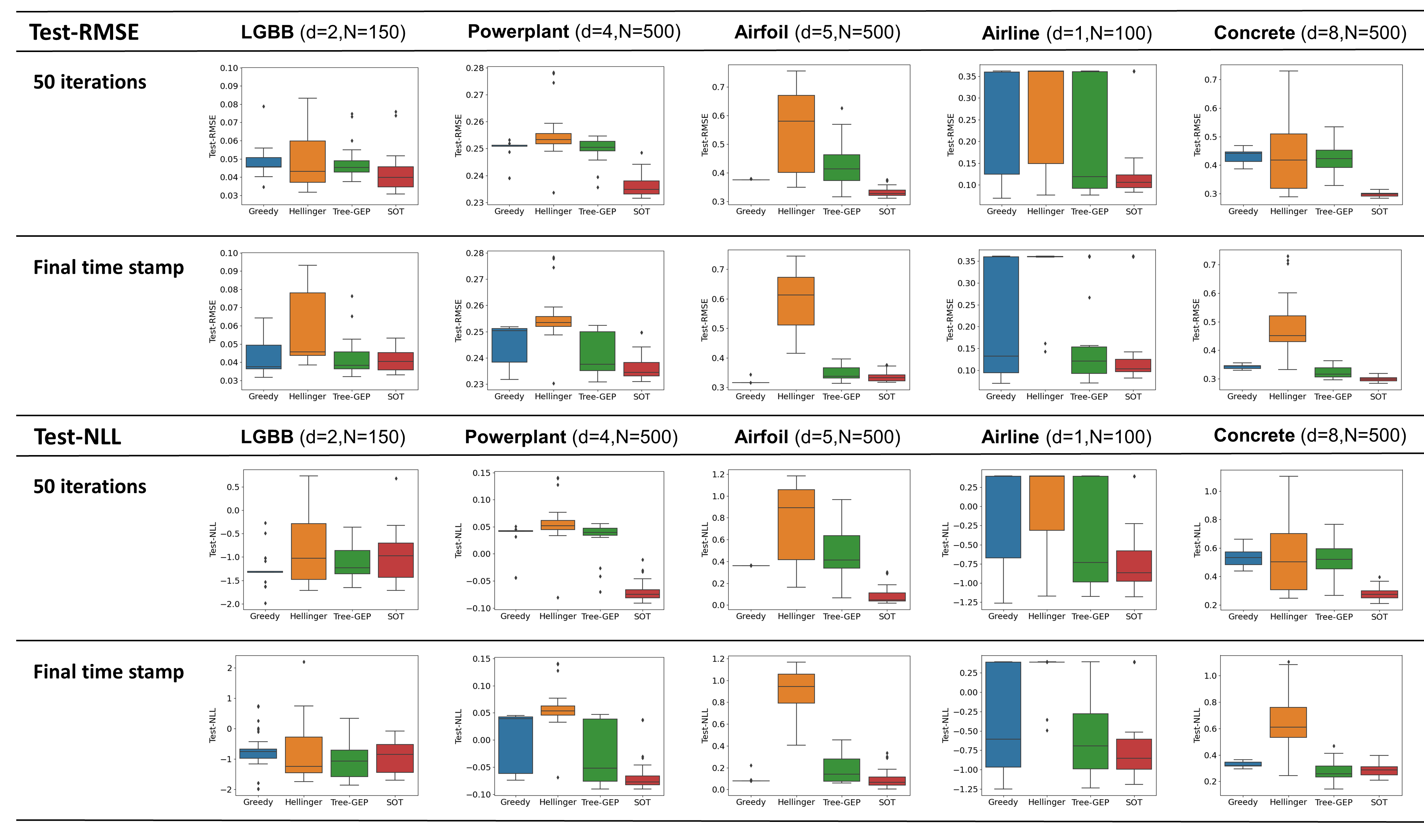}
	\caption{Box-plots of the test-RMSE and test-NLL of the selected hypotheses after 50 iterations and at the final time stamp.}
	\label{fig:testrmseboxplots}
\end{figure}

\paragraph{Test Performance - Box-Plots:} In Figure \ref{fig:testrmseboxplots}, we show box-plots of the RMSE and NLL on the held out test-sets for the selected kernels. We see that on both metrics, our search method also finds a final model that often yields the best test-performance or is among the best models. However, we also note that test performance is mainly a property of the model selection criteria and how well models that maximize that criteria generalize to new data points.
\paragraph{Acquisition Function Optimization vs. Oracle Evaluation Ratio:} In Table \ref{aqvsoracletable}, the ratio $\frac{t_{\mathrm{Acquisition}}}{t_{\mathrm{Oracle}}}$ between the CPU-time needed for optimization of the acquisition function and the CPU-time for evaluating the oracle (calculating the log-evidence) is shown for the SOT kernel-kernel and the Hellinger kernel-kernel \cite{BOMS}. As described in the previous section, the approach in \cite{BOMS} uses a different kind of acquisition function optimization, that has fewer evaluations of the acquisition function than the evolutionary algorithm we use. Nevertheless, we are magnitudes of orders faster. This can be understood more clearly when looking at the raw kernel-kernel evaluation times in Figure \ref{fig:computationtimesummary}. A faster acqusition to oracle time ratio in the end results in a faster search procedure, measured over CPU-Time (as we show in our main results in Figure \ref{fig:mainresults}), which is the metric we are interested in.  

\begin{table}[b]
	\centering
	\caption{Acquisition to oracle time ratio. }
	\begin{tabular}{lll}
		\toprule
		{Dataset} & Hellinger & SOT (ours)\\
		\midrule
		$\mathrm{Airfoil} (N=500)$     &                                            18.77 &            0.25 \\
		$\mathrm{Airline} (N=100)$ &                                                54.21 &            0.825 \\
		$\mathrm{LGBB} (N=150)$       &                                             42.63 &            0.52  \\
		$\mathrm{Powerplant} (N=500)$ &                                            16.51 &            0.278  \\
		$\mathrm{Concrete} (N=500)$ &												16.81	&0.213\\
		\bottomrule
	\end{tabular}
	\label{aqvsoracletable}
	
\end{table}

\paragraph{kNN for Meta Prediction:} For the k-nearast-neighbour approach that was used in Table \ref{rmse_table_space1} we employ the same principle as in \cite{BOMS}
who also consider kNN for comparision. The set $\mathbb{K}_\mathrm{complete}$ consists of kernel expressions which form a directed graph where two neighbour nodes are one grammar operation apart from each other. For each test expression $\tilde{k}\in\mathbb{K}_{\mathrm{test}}$, we search in this graph the k expressions $\tilde{k}_{1},\dots,\tilde{k}_{k}\in\mathbb{K}_{\mathrm{train}}$ with shortest path in this directed graph. The prediction of the log-evidence value of $\tilde{k}$ is the average of the log-evidence values of $\tilde{k}_{1},\dots,\tilde{k}_{k}$. The number $k$ of neighbours is determined via cross validation.

\paragraph{Runs and Time Stamps:} In Figure \ref{fig:mainresults}, all methods were repeated over 30 seeds, where each seed corresponds to a different initial dataset for the both BO methods and TreeGEP and a different ordering of neighbor evaluations in greedy search.  All runs have different run-times. The reasons for this is that the oracle evaluation times differ depending on the kernel that is evaluated. The log-evidence can be calculated faster for smaller hypothesis. The final time stamps in Figure \ref{fig:mainresults} are therefore determined by the shortest run of our method, as we only have log-evidence values for all runs/seeds up to that time point. In rare cases, a run can be interrupted, in case the Laplace approximation returns NaNs. This can happen due to numerical instabilities in the Cholesky decomposition and happened independently of the search method. We filtered out these runs.

\section{Further Experiments}
\label{section:further_experiments}
\begin{figure}[t]
	\centering
	\includegraphics[width=0.99\linewidth]{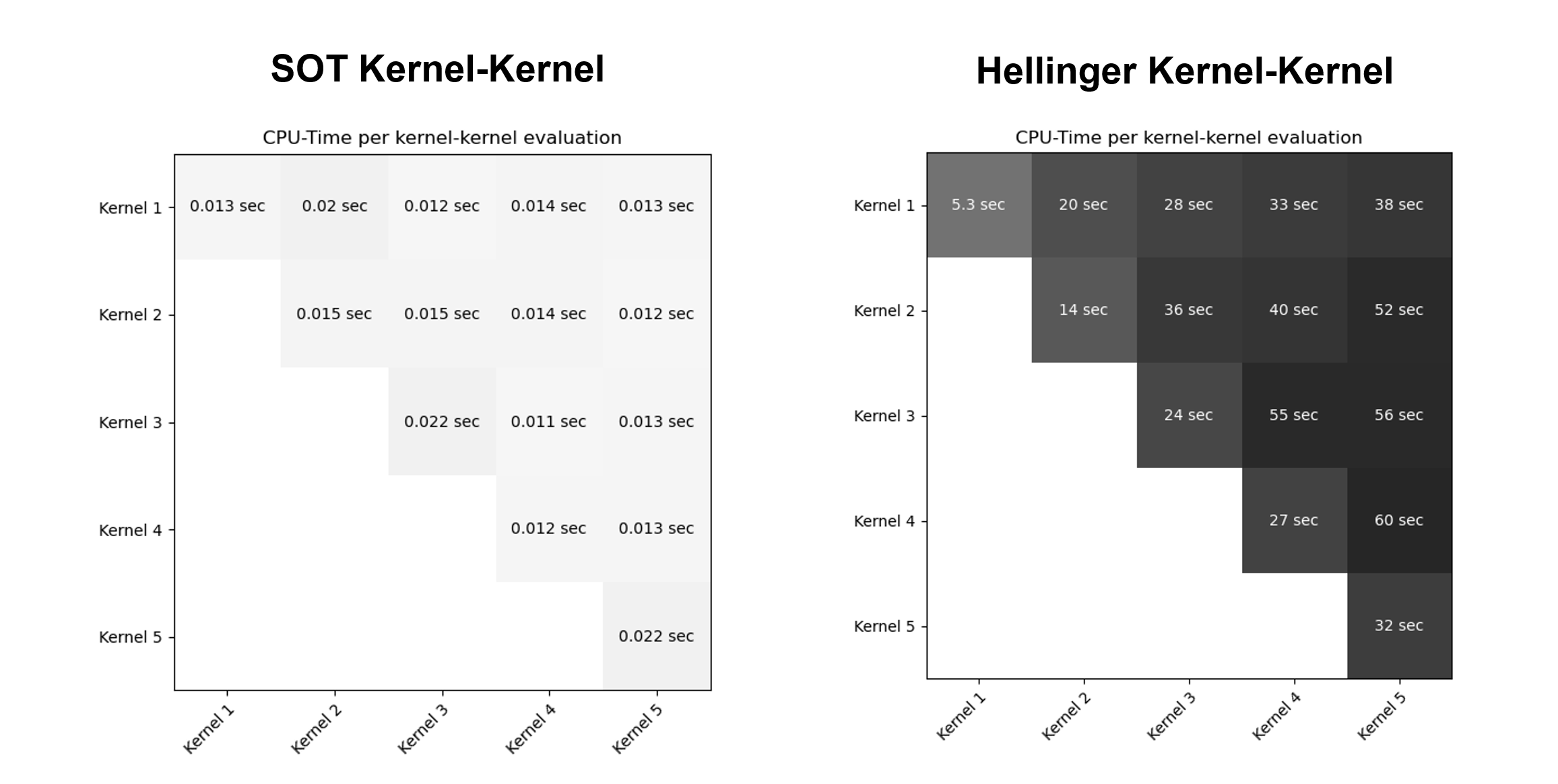}
	\caption{CPU-time for single kernel-kernel evaluations $K(k_{1},k_{2})$ for the SOT kernel-kernel and the Hellinger kernel-kernel.}
	\label{fig:computationtimesummary}
\end{figure}

\paragraph{Computational Time for Single Kernel-Kernel Evaluations:}
In Figure \ref{fig:computationtimesummary}, we show CPU-time for single kernel-kernel evaluations $K(k_{1},k_{2})$ for the SOT kernel-kernel and the Hellinger kernel-kernel, each evaluated on five kernels generated from the kernel grammar (search space was the same as used for the $\mathrm{LGBB}$ dataset). We selected the five kernels with increasing numbers of base kernels and, thus, kernel parameters, as this could affect the computation time (see Table \ref{cputable}). Our kernel-kernel can be evaluated orders of magnitudes faster, which also explains the smaller acquisition function optimization times.
\begin{table}[t]
	\centering
	\caption{Summary of kernels used in Figure \ref{fig:computationtimesummary}. }
	\begin{tabular}{llllll}
		\midrule
		Kernel-Index    & 1 & 2 & 3 & 4  & 5  \\
		\midrule
		\# Parameters   & 2~ & 5~ & 8~ & 10 & 12 \\
		\# Base Kernels & 1 & 2 & 3 & 4  & 5 \\
		\midrule
	\end{tabular}
	\label{cputable}
\end{table}

\begin{figure}
	\centering
	\includegraphics[width=0.9\linewidth]{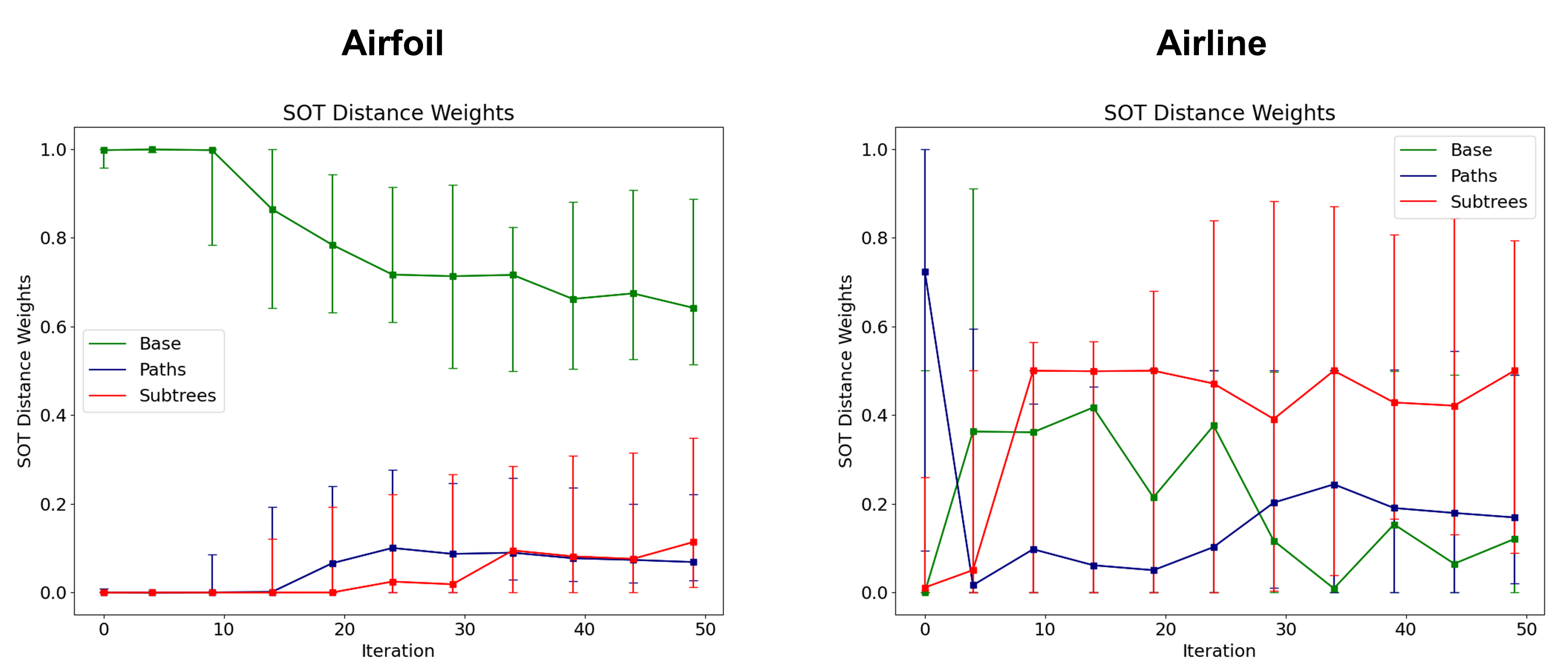}
	\caption{Distance weights of the SOT kernel-kernel over the BO iterations.}
	\label{fig:kernelhps}
\end{figure}

\paragraph{Kernel-Kernel Hyperparameters:}
In Figure \ref{fig:kernelhps}, we show the values of the distance weights for the three OT metrics $W_{\tilde{d}}(\omega_{1,\mathrm{base}},\omega_{2,\mathrm{base}})$ (on $\mathrm{Airfoil}$ summed over dimensions), $W_{\tilde{d}}(\omega_{1,\mathrm{paths}},\omega_{2,\mathrm{paths}})$ and $W_{\tilde{d}}(\omega_{1,\mathrm{subtrees}},\omega_{2,\mathrm{subtrees}})$ over the BO iterations on $\mathrm{Airline}$, and $\mathrm{Airfoil}$. The weights were fitted via marginal likelihood maximization as described in \ref{section:marginal_likelihood}. First, we observe that all OT metrics are used. Secondly, we see that it is dependent  on the dataset which OT metric is used primarily. For example, on $\mathrm{Airline}$ the $\mathrm{Subtree}$ features are the most important ones (according to the marginal likelihood maximization) whereas on $\mathrm{Airfoil}$ the $\mathrm{Base}$ features obtain the largest weights. 
\paragraph{Interpretability of Final Hypothesis:}
The kernel grammar is used in a line of work called Automatic Statistician \cite{CKS,AutomaticStatistician}. In particular \cite{AutomaticStatistician}, utilize the kernel grammar to automatically generate natural language description of the data from the selected hypothesis (they also employ greedy search). Here, we show a configuration of our algorithm where the method of \cite{AutomaticStatistician} can be applied to the final hypothesis of our model selection procedure and can give good descriptions of the dataset. As done in \cite{AutomaticStatistician}, we drop the rational quadratic kernel from the search space, as this captures small and long range correlations at the same time, which can also be modelled with separate squared exponential kernels (see \cite{AutomaticStatistician}). Furthermore, we take fewer steps in the acquisition function optimizer such that fewer base kernels are maximally possible in the final hypothesis. This renders the final hypothesis smaller and easier to interpret, as fewer components needs to be described. We show two example hypotheses that were selected for the $\mathrm{Airline}$ dataset using the described search space and four steps in the acquisition function optimizer. We used the principles presented in \cite{AutomaticStatistician} to simplify the expression and to generate the sentences. 

\textbf{Example 1:}
\begin{align*}
&\mathrm{LIN}+\mathrm{SE} + \mathrm{LIN} \times \mathrm{SE}+ \mathrm{PER} \times \mathrm{LIN} \times \mathrm{SE} \\
\end{align*}
\textbf{Desciption - Example 1:} The data can be described as a sum of:
\begin{enumerate}
	\item A linearly increasing function
	\item A smooth function
	\item A smooth function with increasing variation
	\item An approx. periodic function with linearly increasing amplitude
\end{enumerate}
\textbf{Example 2:}
\begin{align*}
&\mathrm{SE} + 2 \, \mathrm{LIN}^{2}\times \mathrm{PER} + 2\,\mathrm{SE}\times \mathrm{PER}\times \mathrm{LIN}\\
\end{align*}
\textbf{Desciption - Example 2:} The data can be described as a sum of:
\begin{enumerate}
	\item A smooth function
	\item Two approx. periodic functions with linearly increasing amplitude
	\item Two periodic functions with quadratically increasing amplitude
\end{enumerate}

\paragraph{Comparision with RBF Kernel:}
In Table \ref{fklTest} the test-set results (RMSE and NLL) of our proposed kernel search method is shown after 50 iterations compared to test values of a standard RBF kernel. We note that Table \ref{fklTest} shows different NLL values as Table \ref{rmseTest} as it shows values at last iteration and not at last time stamp (see Appendix \ref{section:experimental_details} for details). In terms of NLL, the RBF kernel shows competitive performance on the \textrm{LGBB} and \textrm{Concrete} dataset. However, considering the RMSE and NLL values on the other three datasets, it appears that kernel selection in general seems to be very important.

\paragraph{Comparision with Nonparametric Kernel Learning Methods:}
In Table \ref{fklTest} we also compare with the nonparametric kernel learning method, presented in  \cite{FKL}, called Function Kernel Learning (FKL). FKL places a GP prior on the spectral density of the kernel - thus utilizing a nonparametric approach to kernel learning/selection. This results in a prior over spectral-mixture kernels, thus a prior over a fixed but highly flexible kernel structure. The results in Table \ref{fklTest} might be an indication that search over a discrete set of structural kernels might be beneficial, compared to placing a prior over a very flexible, but fixed kernel family.

\begin{table}[t]
	\centering
	\caption{RMSE and predictive NLL values on the respective test-sets after 50 iterations of our proposed search method + the RMSE/NLL values of FKL \cite{FKL} and a standard RBF kernel. SOT values are marked bold if they are not significantly different from the best value (FKL and RBF are point evaluations) according to a t-test ($\alpha=0.05$).}
	\label{fklTest}
	\begin{tabular}{lcccccccl}
		\toprule
		Dataset &  SOT (ours) & FKL & RBF &  SOT (ours) & FKL & RBF\\
		\midrule
		\multicolumn{6}{l}{~~~~~~~~~~~~~~~~~~~~~~~~~~~~~~~~~~~~~~~~~~~~~~\textbf{RMSE} ~~~~~~~~~~~~~~~~~~~~~~~~~~~~~~~~~~~~~~~~~~~~~~~~~~~\textbf{NLL}}  \\
		\cmidrule(r){2-4} \cmidrule(r){5-7}
		$\mathrm{Airline}$  & \textbf{0.1335} (0.079)  & 0.3614 & 0.3598 & \textbf{-0.7069} (0.442)  & 0.4712 & ~0.3978\\
		$\mathrm{LGBB}$     & \textbf{0.0422} (0.011)  & 0.1296 &0.0740  & \textbf{-0.9762} (0.517)  & 0.0200 &\textbf{-1.0913}\\
		$\mathrm{Powerplant}$  & \textbf{0.2362} (0.004) & 0.2532 & 0.2507 & \textbf{-0.0693} (0.019) & 0.1755 & ~0.0419\\
		$\mathrm{Airfoil}$  & \textbf{0.3334} (0.017)  & 0.4356 &0.4075 & ~\textbf{0.0855} (0.081)  & 0.6469 &~0.2985\\
		$\mathrm{Concrete}$  & \textbf{0.2980} (0.008) &  0.3230 &0.3617 & ~\textbf{0.2787} (0.044)  &  4.2150 &~\textbf{0.2743}\\
		\bottomrule
	\end{tabular}
\end{table}

\begin{figure}[h]
	\centering
	\includegraphics[width=0.94\linewidth]{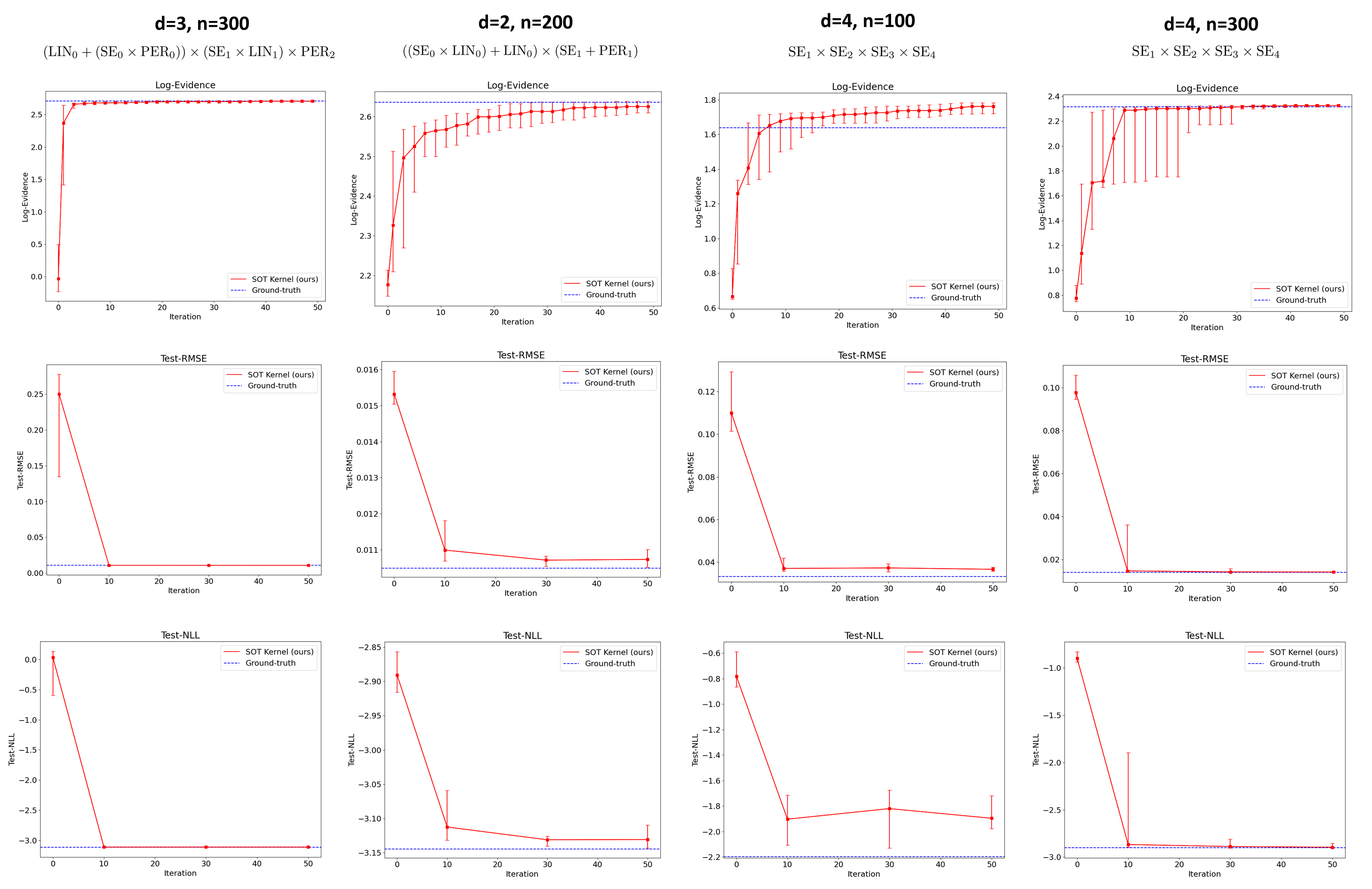}
	\caption{Learning-curves of our kernel selection method (over iterations) for simulated datasets. Upper plot shows log-evidence values, lower plot shows test-set RMSE and NLL values. Blue lines mark log-evidence/test values of the ground-truth kernel, from which the dataset was generated. }
	\label{fig:gtkernelcomparision}
\end{figure}

\paragraph{Experiments on Simulated Data \& Type-3 Maximum Likelihood Overfitting:}
In Figure \ref{fig:gtkernelcomparision}, we show experimental results on simulated data.
 Here, we draw $n$ datapoints from a GP prior with a given kernel structure and employ our kernel selection method on that dataset (likelihood noise was 0.01 for all datasets). First, we observe that we reach (almost) the same log-evidence values as the ground truth kernel within 50 iterations. For the RBF ground-truth kernel we simulated one big dataset and used $n=100$ and $n=300$ for model selection in the third and fourth plot. For the smaller dataset, we observe that we find kernels that have even higher log-evidence values than the ground-truth kernel. Considering the test-set NLL we also observe a small increase in the NLL from iteration 10 to 30, indicating a small overfitting. We think that in particular for smaller datasets maximizing the log-evidence over kernel structures can also lead to overfitting. However, this might not be surprising as the same was observed for type-2 maximum likelihood in GP's (see. \cite{PromisesAndPitfallsofDKL}) whereas maximizing the log-evidence might be interpreted as type-3 maximum likelihood. Possible methods to avoid overfitting could be to use a different model selection criterion such as the Bayesian information criterion or cross-validation error, or to use a smaller search space.

\end{document}